\documentclass[10pt,twocolumn,letterpaper]{article}

\usepackage[pagenumbers]{cvpr}

\usepackage{tabularx}
\usepackage{multirow}
\usepackage{enumitem}
\usepackage{pifont}
\usepackage{bbm}
\usepackage{overpic}
\usepackage{subcaption}
\usepackage{amsfonts}
\usepackage{booktabs}
\usepackage{nicefrac}
\usepackage{wrapfig}
\usepackage{microtype}
\usepackage[many]{tcolorbox}
\usepackage{floatflt}
\usepackage{textcomp}
\usepackage[export]{adjustbox}
\usepackage[accsupp]{axessibility}

\newcommand\doublecheckmark{\textcolor{black}{\checkmark\kern-0.45em\checkmark}}
\newcommand{\inlineColorbox}[2]{\begingroup\setlength{\fboxsep}{1pt}\colorbox{#1}{\hspace*{2pt}\vphantom{Ay}#2\hspace*{2pt}}\endgroup}
\definecolor{DrawioBlue}{RGB}{218,232,252}
\definecolor{DrawioOrange}{RGB}{255,230,204}
\definecolor{DrawioGreen1}{RGB}{204,255,204}
\definecolor{DrawioGreen}{RGB}{213,232,212}
\definecolor{DrawioPurple}{RGB}{225,213,231}
\definecolor{DrawioRed}{RGB}{248,206,204}
\definecolor{DrawioYellow}{RGB}{255, 242, 207}
\definecolor{lightgray}{RGB}{225,225,225}
\definecolor{DrawioDarkPurple}{HTML}{9673A6}
\definecolor{DrawioDarkRed}{HTML}{B85450}
\definecolor{DrawioDarkGreen}{HTML}{009900}
\definecolor{DrawioDarkBlue}{HTML}{6C8EBF}

\definecolor{cvprblue}{rgb}{0.21,0.49,0.74}
\usepackage[pagebackref,breaklinks,colorlinks,allcolors=cvprblue]{hyperref}

\def\confName{CVPR}
\def\confYear{2025}

\title{Compositional Caching for Training-free Open-vocabulary Attribute Detection}

\author{Marco Garosi\textsuperscript{1}\quad
Alessandro Conti\textsuperscript{1}\quad
Gaowen Liu\textsuperscript{2}\quad
Elisa Ricci\textsuperscript{1,3}\quad
Massimiliano Mancini\textsuperscript{1} \\
{\small \textsuperscript{1}University of Trento\quad\textsuperscript{2}Cisco Research\quad\textsuperscript{3}{Fondazione Bruno Kessler}} \\
{\small \url{https://comca-attributes.github.io}}
}

\newcommand{\ours}{\textsc{ComCa}\xspace}
\newcommand{\oursFull}{Compositional Caching\xspace}

\begin{document}
\maketitle
\begin{abstract}
Attribute detection is crucial for many computer vision tasks, as it enables systems to describe properties such as color, texture, and material.
Current approaches often rely on labor-intensive annotation processes which are inherently limited: objects can be described at an arbitrary level of detail (\eg, color vs. color shades), leading to ambiguities when the annotators are not instructed carefully.
Furthermore, they operate within a predefined set of attributes, reducing scalability and adaptability to unforeseen downstream applications.
We present \textsc{Compositional Caching} (\ours), a training-free method for open-vocabulary attribute detection that overcomes these constraints.
\ours requires only the list of target attributes and objects as input, using them to populate an auxiliary cache of images by leveraging web-scale databases and Large Language Models to determine attribute-object compatibility.
To account for the compositional nature of attributes, cache images receive soft attribute labels. Those are aggregated at inference time based on the similarity between the input and cache images, refining the predictions of underlying Vision-Language Models (VLMs).
Importantly, our approach is model-agnostic, compatible with various VLMs.
Experiments on public datasets demonstrate that \ours significantly outperforms zero-shot and cache-based baselines, competing with recent training-based methods, proving that a carefully designed training-free approach can successfully address open-vocabulary attribute~detection.
\end{abstract}

\section{Introduction}
\begin{figure}[t!]
\centering
\includegraphics[width=\columnwidth]{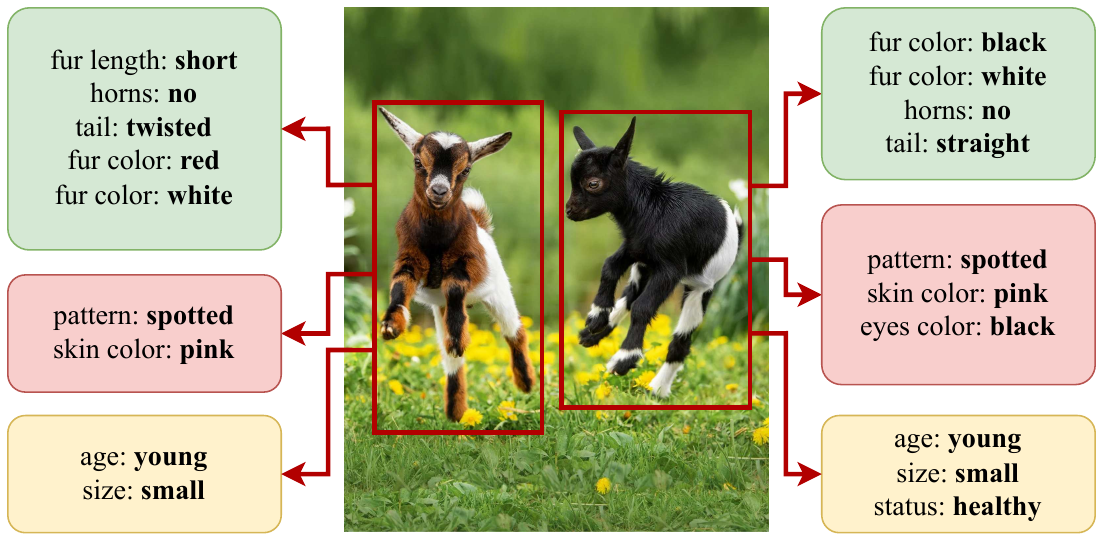}
\caption{
    \textbf{Example of attribute annotations.}
    Attribute annotations are often: \inlineColorbox{DrawioGreen}{sparse}, as they are not consistent across samples; \inlineColorbox{DrawioRed}{incomplete}, as not all attributes are annotated; \inlineColorbox{DrawioYellow}{ambiguous}, as they can be subjective or miss a frame of reference.
    This makes open-vocabulary attribute detection challenging.
    Differently from previous works~\cite{bravo2023ovad,chen2023ovarnet}, we do not rely on such annotations and propose \ours, a training-free approach requiring no supervision.
}
\label{fig:teaser}
\vspace{-5mm}
\end{figure}

Attributes shape our perception and interactions with the environment.
Let us consider the example of \cref{fig:teaser}.
If our aim is to identify the animal, we may focus on its fur color, size, the presence of horns and the tail. Similarly, if we want to analyze its health status, we may focus on its age, fur patterns and shades, and its skin.
This level of analysis requires models with fine-grained understanding of the visual input, which is usually obtained by training a model on datasets that annotate both objects and their corresponding attributes~\cite{patterson2016coco, pham2022improving}.
However, collecting attribute annotations is both time-consuming and prone to ambiguities, especially when annotators lack clear instructions. Additionally, attribute annotations are often tailored to specific datasets and are ambiguous.
For instance, in \cref{fig:teaser}, one may list ``brown'' as a color for the left goat, but others may argue that it is a red-ish shade of brown.
A simple solution is to provide annotators with a predefined list of attributes; however,
(i) the list may not cover all possible annotations for a property (\eg, all colors) and
(ii) it limits models trained on it to the given list itself, making them unable to recognize new attributes.

Recent approaches overcome some of the challenges by developing open-vocabulary attribute detection models~\cite{bravo2023ovad, chen2023ovarnet, guo2023lowa} that detect attributes specified in natural language at inference time.
While these approaches generalize beyond the set of seen attributes and objects, their performance is still biased toward the specific dataset they are trained on.
Indeed, training-based approaches may show significant degradation in performance in cross-dataset experiments (see \cref{fig:cross_dataset}).

In this work, we overcome potentially ambiguous and/or limited attribute annotations by introducing \textsc{\textbf{Com}positional} \textsc{\textbf{Ca}ching} (\textbf{\ours}), a \textit{training-free} open-vocabulary attribute detection model.
\ours employs a Vision-Language Model (VLM)~\cite{radford2021clip} and refines its predictions by utilizing an auxiliary cache that contains images together with \textit{automatically estimated} attribute labels.

Caching for refining CLIP predictions has been introduced in the context of image classification~\cite{zhang2022tip}.
It requires storing images, either generated or retrieved~\cite{udandarao2023sus, zhang2022tip}, for each of the classes of interests, together with a corresponding one-hot categorical label.
The similarity of the input image with each of the examples is used to weight the labels in the cache and refine the model's prediction.
While this approach is effective for object classification, it is prohibitively expensive in the context of attribute detection due to the much larger output space, comprising all possible combinations of attributes and objects.
\ours achieves scalability in caching by following two key compositional principles: (i) not all attributes are universally applicable to all objects, and (ii) every object inherently possesses multiple attributes.

The first principle is implemented in two steps: attribute-object compatibility estimation and cache pair sampling.
In the first phase, we estimate
the probability of an attribute appearing with an object using the statistics of available databases (\eg, CC12M~\cite{changpinyo2021cc12m}) and Large Language Model (LLM) queries.
Next, we populate the cache for a given attribute by sampling compatible objects from the estimated distribution and retrieving relevant images from a database.
The second principle is implemented by self-labeling the cached elements. Initially, each sample is associated with the single attribute it was sampled for.
Since this does not account for other attributes present in the image, we extend attribute labels through soft-labeling, \ie, comparing cache images to text embeddings of each attribute of interest.
These scores modify a cache element's contribution to each target attribute, considering their implicit composition.
We evaluate \ours on two publicly available benchmarks, OVAD~\cite{bravo2023ovad} and VAW~\cite{pham2021vaw}. \ours demonstrates significant improvements over training-free baselines and is competitive with recent training-based approaches, while remaining unaffected by the biases of specific training sets. Notably, \ours provides a consistent boost in performance regardless of the VLM pre-training and architecture.

In summary, our contributions are: 
\textcolor{teal}{\ding{172}} We propose \ours, an approach specifically designed for training-free open-vocabulary attribute detection. \ours exploits the compositional nature of objects and attributes to build an auxiliary cache that is used as an anchor to refine the model's predictions.
\textcolor{teal}{\ding{173}} We make caching scalable by extracting priors from databases and LLMs on which objects are associated with each attribute, using them to populate the cache. Soft-labeling further accounts for the multiplicity of attributes within an image.
\textcolor{teal}{\ding{174}} We empirically show that \ours outperforms all the cache-based baselines and achieves competitive performance against costly training-based methods, demonstrating the practical feasibility of training-free open vocabulary attribute detection.

\section{Related work}

\begin{figure*}[t!]
\makebox[\linewidth][c]{\includegraphics[width=1.0\textwidth]{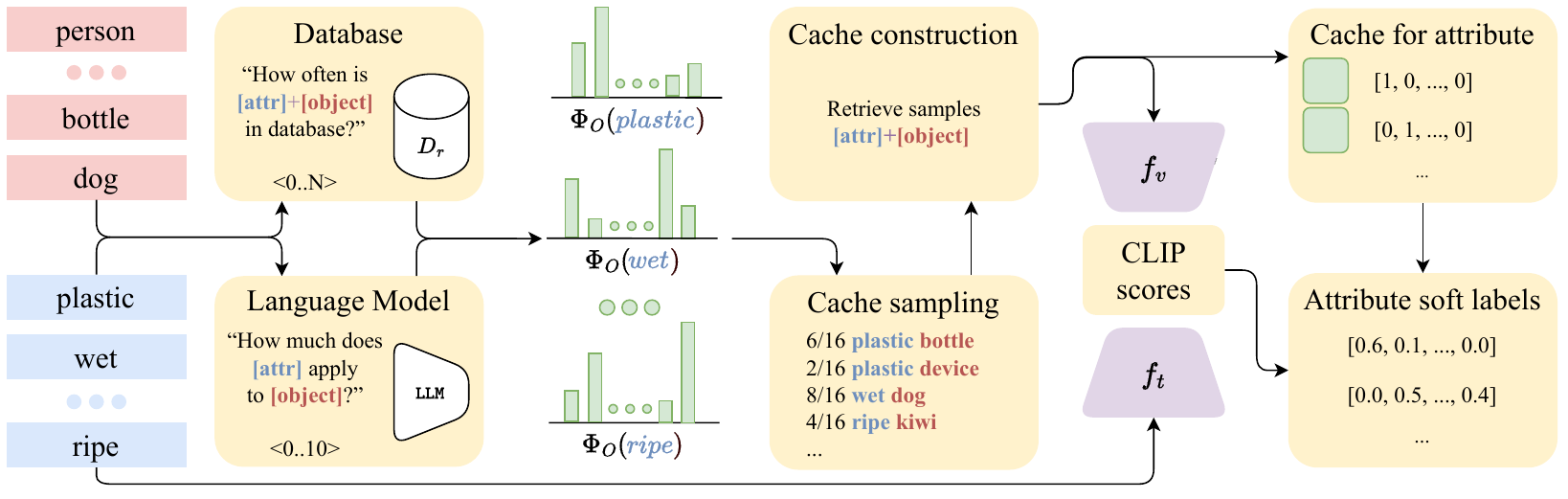}}
\caption{
\textbf{\ours's cache construction.} Given a list of \inlineColorbox{DrawioBlue}{attributes} and \inlineColorbox{DrawioRed}{objects}, we compute their compatibility from a large \inlineColorbox{DrawioYellow}{database}
$D_r$ and with an  \inlineColorbox{DrawioYellow}{LLM}.
The scores are merged and normalized to obtain the \inlineColorbox{DrawioGreen}{compatibility distribution}, from which we \inlineColorbox{DrawioYellow}{sample cache entries} and \inlineColorbox{DrawioYellow}{construct the cache}.
We enrich the latter with soft labels from the VLM-based similarity between cache images and \inlineColorbox{DrawioBlue}{attributes}.
}
\label{fig:method}
\vspace{-5mm}
\end{figure*}

\noindent\textbf{Attribute detection.}
Detecting visual attributes is a widely explored task in computer vision~\cite{ferrari2007learning, farhadi2009describing, farhadi2010attribute, berg2010automatic}, supporting various downstream applications such as zero-shot learning~\cite{lampert2009learning, akata2013label, lampert2013attribute, xian2018zero}, relative image comparison~\cite{parikh2011relative, yu2014fine, chen2018compare, siddiquie2011image}, and compositional image retrieval~\cite{hosseinzadeh2020composed, baldrati2023zero, saito2023pic2word, karthik2023vision}.

Recent works have introduced several benchmarks to promote and ease research in this area~\cite{krishna2017visual, patterson2016coco, pham2021vaw, isola2015discovering, pham2022improving, han2017automatic, welinder2010caltech, wu2021fashion, yu2014fine}.
Despite these advances, annotating visual attributes remains a complex and challenging task.
This is due to the wide variability in attributes across objects and contexts, as well as the difficulty of ensuring consistent annotations when multiple annotators are involved.
Moreover, many datasets capture only a limited range of object attributes, limiting their usefulness for developing general-purpose models.

Open-vocabulary attribute detection (OVAD)~\cite{bravo2023ovad, chen2023ovarnet} overcomes some of these limitations by detecting user-specified attributes, typically provided in natural language, even those that have not been observed during training.
This task was first explored in~\cite{bravo2023ovad}, which introduced the OVAD benchmark and method.
OVAD generates supervisory signals for attributes, nouns, and noun phrases by parsing captions and encoding them using CLIP text embeddings~\cite{radford2021clip}.
OvarNet~\cite{chen2023ovarnet} builds on this by fine-tuning CLIP with federated learning and additional supervision from external captions.
LOWA~\cite{guo2023lowa} proposes a three-stage pipeline that aligns visual embeddings to objects, attributes, and free-form text.
ArtVLM~\cite{zhu2024artvlm} uses a generative VLM to obtain attribute probabilities based on the input image.
Although these methods operate in the open-vocabulary setting, they are training-based and require both a lot of annotated data and several computational resources (see \cref{sec:exp}).
\ours significantly differs from these approaches, as we explore for the first time a training-free, cache-based, strategy for attribute detection.

\noindent\textbf{Adapting Vision-Language Models.}
VLMs~\cite{radford2021clip, li2022blip, li2023blip, yu2022coca, liu2024llava} bridge the gap between textual and visual data, showing remarkable zero-shot generalization abilities.
Recent efforts have focused on improving their performance on specific downstream tasks through various strategies, broadly categorized into training-based and training-free methods.

Training-based approaches typically involve tuning text prompts~\cite{yao2023prompt, zhu2023prompt, sohn2023prompt, zhou2022cocoop, jia2022visual}.
Some methods learn prompts directly~\cite{zhou2022coop} or predict them~\cite{zhou2022cocoop} from visual inputs to improve generalization.
Other approaches train adapters to refine the model's predictions~\cite{hu2021lora, gao2024adapter}.
While effective, these methods introduce additional parameters, which can negatively impact the model's zero-shot performance~\cite{zhou2022cocoop}.

Training-free methods take a different approach, such as incorporating additional knowledge into the prompts using natural language~\cite{menon2022visual, novack2023chils, roth2023waffling}, or modifying predictions through attention layers~\cite{guo2023calip} or normalization techniques~\cite{zhou2024test}.
Recent works also employ a cache of images and one-hot labels that represent the target classes~\cite{zhang2022tip, udandarao2023sus, zhu2023not}.
For example, TIP-Adapter~\cite{zhang2022tip} computes the similarity between input images and cached examples, using the associated labels for prediction.
Similarly, SuS-X~\cite{udandarao2023sus} creates a cache automatically, either by using a generative model~\cite{rombach2022high} or retrieving from a large database~\cite{schuhmann2022laion}, with distance-based scoring to weight the elements' contribution.
While our work builds upon the same principles
of TIP-Adapter and SuS-X, \ie, using a cache to refine predictions, it differs in four key aspects, as \ours's cache:
(i) is aware of attribute-object compatibility; (ii) accounts for attribute combinations; (iii) models the compositional nature of attributes and objects; (iv) scales efficiently with the number of attributes.

\section{\oursFull}
\label{sec:method}

In this section, we formally define the open-vocabulary attribute detection problem and describe caching as a tool for VLM adaptation (\cref{sec:problem}). Next, we describe how attribute-object compatibility informs the construction of scalable caches (\cref{sec:caching}). Finally, we detail our process for refining cache labels, taking into account multi-attribute compatibility (\cref{sec:soft-caching}). \cref{fig:method} provides an overview of our approach.

\subsection{Background}
\label{sec:problem}

\noindent\textbf{Problem formulation.} Given an image $x \in \mathcal{X}$ of an object and a list of attributes, our goal is to detect which attributes are present in the image.
Let $\mathcal{A}$ be the set of all existing attributes in natural language. At inference time, the model is provided with a set of $n$ target attributes $A = \{a_1, \cdots, a_n\}$ as input, where $A \subset \mathcal{A}$.
These are associated with $m$ target objects, which we denote as $O = \{o_1, \cdots, o_m\}$.
Our goal is to build a function $f: \mathcal{X} \times \mathcal{A}^n \rightarrow [-1, 1]^{n}$ that maps an image and a list of attributes to a binary vector indicating the presence ($1$) or absence ($-1$) of each attribute.

We obtain $f$ without training, in contrast to previous works that learn $f$ using a dataset ${D}$ of image-label pairs $(x, \mathbf{a})$ with $\mathbf{a} \in [-1,1]^{n}$ being the ground-truth binary vector\footnote{In some problem formulations, the dataset may contain (i) object labels, (ii) bounding boxes, (iii) an ``unknown'' value for some attributes. For simplicity, we omit the latter from the formulation and, following previous works~\cite{pham2021vaw}, we consider the case where objects bounding boxes are given.}.
We achieve this by implementing $f$ using a VLM combined with an auxiliary cache to refine its predictions.

\noindent\textbf{Adapting VLMs with caching.}
Following previous works~\cite{bravo2023ovad, chen2023ovarnet}, we implement $f$ with a VLM. While we primarily consider CLIP~\cite{radford2021clip} to describe our method and perform our evaluation, the proposed framework can integrate several other VLMs (see \cref{table:main_results}).
The function $f$ has three elements: a visual encoder $f_{v}$, a text encoder $f_{t}$, and a prediction function.
$f_v$ and $f_t$ map images in $\mathcal{X}$ and natural language in $\mathcal{T}$, respectively, to a shared $d$-dimensional multimodal space $\mathcal{Z} \in \mathbb{R}^d$, \ie, $f_v: \mathcal{X} \rightarrow \mathcal{Z}$ and $f_t: \mathcal{T} \rightarrow \mathcal{Z}$.
Next, the prediction function maps embeddings from the multimodal space to the final scores.
A standard way for scoring an image $x$ against a class $c$ is by computing the cosine similarity between their embeddings:\footnote{For simplicity, we omit the presence of prompts such as \texttt{"a photo of"} when defining the text input.}
\begin{equation}
    \label{eq:clip}
    f_\mathtt{CLIP}(x,c) = \frac{f_v(x)^{\intercal} f_t(c)}{||f_v(x)|| \cdot||f_t(c)||}. 
\end{equation}

While effective, $f$ can be improved without any training by using an auxiliary cache that stores examples of the classes of interest~\cite{orhan2018simple, zhang2022tip, udandarao2023sus}. Given a cache $\mathcal{C}=\{x_1^1, \cdots, x^K_C\}$ that contains $K$ samples of $C$ classes,
we can define the cache-based prediction for a class $c\in C$ as:
\begin{equation}
    \label{eq:cache-tip}
    f_\mathcal{C}(x,c) = \sum_{i=1}^{K} \eta_\mathcal{C} \left(\frac{f_v(x)^{\intercal} f_v(x^i_c)}{||f_v(x)|| \cdot ||f_v(x^i_c)||} \right),
\end{equation}
where $\eta_\mathcal{C}(z) = \exp(1-\beta(-z))$ is a normalization function regulated by the hyperparameter $\beta$~\cite{zhang2022tip}.

The formulation in \cref{eq:cache-tip} has a main issue when applied to open-vocabulary attribute detection: the cache construction. 
On one hand, a cache considering all possible attribute labels is semantically complete but requires a prohibitive storage size of $K \cdot n!$ images, \eg, $K \cdot 117!$ for OVAD~\cite{bravo2023ovad}. On the other hand, while storing $K$ positive examples per attribute has low memory requirements, it ignores the multi-attribute nature of images (\eg, \textit{shape}, \textit{size}, \textit{color}, etc. can co-exist in the same sample), potentially leading to incorrect predictions (\eg, a high score because the \textit{color} matches but the cache label refers to the \textit{shape}). Moreover, attributes may appear differently across objects (\eg, \textit{wet car} vs \textit{wet dog}) and selecting a random subset of them may lead to a cache representative only of a few cases.
In the following, we address these issues by focusing on the compatibility between objects and multiple attributes in the real world.

\subsection{Compatibility for scalable caching}
\label{sec:caching}

Naively constructing the cache would lead to many problems, related both to labels' semantic and memory requirements (\cref{sec:problem}).
In this section, we present a methodology that (i) has the same memory requirements as storing only $K$ examples per attribute and (ii) assigns multiple attribute labels per example, taking into account attribute-object compatibility.

\noindent\textbf{Scalable caching.}
To obtain a cache $\mathcal{C}$ that is scalable with the number of attributes and accounts for object-specific information, we exploit the natural distributions of attributes across objects.
In particular, let us define a function $\varphi$ that,
given an attribute $a \in \mathcal{A}$ and an object $o \in \mathcal{O}$,
returns their compatibility value as a scalar, \ie, $\varphi: \mathcal{A} \times \mathcal{O} \rightarrow \mathbb{R}$.

Considering an attribute $a$ and an object $o$, the score $\varphi(a, o)$ is proportional to the probability that $o$ exhibits $a$.
Thus, we can use the scores $\Phi_O(a) = [\varphi(a, o_1), \ldots, \varphi(a, o_m)]$ as a proxy for the probability distribution of $a$ co-occurring with the elements of
$O$.
We then assign each of the $K$ elements (\ie, shots) of the attribute $a$ to one object $o$, sampling the object from $\Phi_O(a)$:
\begin{equation}
    \label{eq:sampling}
    \hat{O}^K_a = \{\hat{o}^a_1, \ldots, \hat{o}^a_K\},\;\; \text{where} \;\;\hat{o}^a_i \sim \Tilde{\Phi}_O(a),
\end{equation}
where $\Tilde{\Phi}_O(a)$ is the normalized version of ${\Phi}_O(a)$,
\ie, $\Tilde{\Phi}_O(a) = \Phi_O(a) / \sum \Phi_O(a)$.
The set $\hat{O}^K_a$ allows populating the cache with relevant objects and attributes.
Let us denote a generic function mapping a textual query to an image as $\mathtt{T2I}: \mathcal{T} \rightarrow \mathcal{X}$. The elements of the cache are:
\begin{equation}
    \label{eq:cache}
    \begin{split}
    \mathcal{C}_\mathtt{H}
        &= \{x_{a_1}^1, \ldots, x_{a_m}^K\} \\
        &= \{\mathtt{T2I}(\pi_g \circ a \circ o^a_1), \ldots, \mathtt{T2I}(\pi_g \circ a \circ o^a_K)\},
    \end{split}
\end{equation}
where $\pi_g$ is a contextual prompt for the image (\eg, \texttt{"a photo of"}).
Following SuS-X~\cite{udandarao2023sus}, we implement $\mathtt{T2I}$ via \textit{retrieval}.
Specifically, we retrieve images from a database by scoring them against textual queries with the VLM (\eg, CLIP). While we focus on retrieval, \ours also supports the implementation of $\mathtt{T2I}$ via a text-to-image generative model, \eg, by feeding textual queries to Stable Diffusion~\cite{rombach2022high}.
We refer to the \textit{Supp. Mat.} for more details.

\noindent\textbf{Estimating attribute-object compatibility.}
The key element that enables scalable caching is the function $\varphi$ that estimates the viability of attribute-object compositions.
We implement it by merging two scoring strategies.

The first uses a generic large-scale database $D_r$ containing image-caption pairs.
Using the captions $t \in D_r$, we can estimate the database score as the number of co-occurrences of attribute $a$ and object $o$, \ie:
\begin{equation}
    \label{eq:dataset-stat}
   \varphi_\mathtt{db}(a,o) = \sum_{(x,t)\in D_r} \mathbbm{1}(o \subset t \land a \subset t),
\end{equation}
where $\mathbbm{1}(z)$ is an indicator function whose value is $1$ whenever the condition $z$ is true, and $0$ otherwise.
 
While this process exploits real data, it may be biased on the specific captions and/or data collection process.
For instance, captions rarely describe objects in their default state, such as a \textit{dry dog}.
Therefore, a second scoring strategy exploits the reasoning capabilities of an LLM, querying it to predict an attribute-object compatibility as:
\begin{equation}
    \label{eq:llm-stat}
   \varphi_\mathtt{LLM}(a,o) = \mathtt{LLM}(\pi_c, o, a),
\end{equation}
where $\pi_c$ is a prompt specifying the compatibility task
(see the \textit{Supp. Mat.}).
The objective is to assign a \textit{score} on a specific range to the likelihood of the attribute-object pair.

Taking into account that also LLMs have issues, such as hallucinations and biases~\cite{gallegos2023bias, navigli2023biases}, we implement $\varphi$ by combining database- and LLM-based scoring as:
\begin{equation}
    \label{eq:combine-scores}
    \varphi(a, o) = \varphi_\mathtt{db}(a,o) \cdot \varphi_\mathtt{LLM}(a,o),
\end{equation}
thus computing $\Phi_{O}(a)$ and filling the cache accordingly.

\subsection{Compositional cache refinement}
\label{sec:soft-caching}
After the process in \cref{sec:caching}, we have a cache $\mathcal{C}_\mathtt{H}$ populated with $K$ examples for each attribute
$a\in A$.
However, the cache assigns one \textit{hard} attribute to each element, contrasting with the natural co-occurrence of attributes and leading to potentially wrong associations during inference.
As a simple example, an image of a \textit{large} and \textit{blue} car obtained with the binding (\textit{large}, \textit{car}) will have a positive label only for \textit{large}. As a consequence, if the test image shows a \textit{small blue car}, the cache element will not contribute to the score of \textit{blue} and may erroneously increase the score for the antonym \textit{large}.

\noindent\textbf{Soft caching.}
To remedy to this issue, we refine the label associated to each cache element, exploiting the
VLM capabilities to compute \textit{soft} labels. Formally, the soft label $s_a$ for an attribute $a$ and a cache image $x_c$ is defined as:
\begin{equation}
    \label{eq:cache-soft-init}
         s_a(x_c) = \frac{f_t(a)^{\intercal} f_v(x_c)}{||f_t(a)|| \cdot ||f_v(x_c)||},
\end{equation}
where we compare the text embeddings of the attribute with the visual embeddings of the cache image.
While these scores can directly serve as soft labels, we found them to exhibit low-variance due to the well-known modality-gap~\cite{liang2022mind,udandarao2023sus}.
We overcome this limitation and increase their relevance by defining normalized soft labels $\bar{s}_a$ as:
\begin{equation}
    \label{eq:cache-soft}
         \bar{s}_a(x_c) =  \frac{\exp\left(({s_a(x_c)-\mu_C})/{\sigma_C}\right)}{\sum_{\hat{a}\in A}{\exp\left((s_{\hat{a}}(x)-\mu_C)/\sigma_C\right)}},
\end{equation}
where $\mu_C$ and $\sigma_C$ are respectively the mean and standard deviation of the similarity between any attribute with any element of the cache.
The first normalization expands the range of values of the similarity scores $s_a$, while the softmax ensures that all samples have an equal contribution to the final prediction.
In practice, we found the latter step to be important only on datasets with many attributes (\eg, VAW).
We hypothesize this is due to differences in the output space cardinality, requiring different softmax scaling as in \cite{zhang2019adacos}.

These scores are mixed with the original hard labels to obtain the final label $\hat{s}_a(x_c) =(1-\alpha)\cdot\mathbbm{1}(a=c) +  \alpha\cdot\bar{s}_a(x_c)$ where $\alpha$ is a trade-off hyperparameter. With the soft labels $\hat{s}_{a}$, we can perform the cache-based prediction as: 
\begin{equation}
\label{eq:our-cache-scoring}
    f^\mathtt{\ours}_\mathcal{C}(x,a) = \sum_{x_c}^{\mathcal{C}} \eta_\mathcal{C} \left(  \hat{s}_a(x_c) \frac{f_v(x)^{\intercal} f_v(x_c)}{||f_v(x)|| \cdot ||f_v(x_c)||} \right).
\end{equation}
In contrast to \cref{eq:cache-tip}, here we consider each cache element $x_c$ as a potential contributor for attribute $a$, even if $a \neq c$.
Implicitly, this allows us to include an arbitrary number of (positive) attribute labels within each element of the cache.

\noindent\textbf{Final inference.}
Finally, the cache-based scoring defined in \cref{eq:our-cache-scoring} is merged with the standard CLIP-based scores to obtain the final predictions:
\begin{equation}
    \label{eq:final-score}
    f(x,a) = \eta_A\left({\lambda} \cdot  f^\mathtt{\ours}_\mathcal{C}(x,a) + f^\mathtt{CLIP}_\mathcal{C}(x,a)\right),
\end{equation}
where $\lambda$ is an hyperparameter controlling the balance between the two and $\eta_A(z) = \texttt{softmax}_A(z/\max(z))$, moving from simple scalar values to a distribution over the attributes space
and leading to more stable scores.
Note that \cref{eq:final-score} contains \textit{no} element which needs training or labeled data for the specific attributes of interest. Given the target sets of attributes $A$ and objects $O$, \ours automatically re-generates the cache to update the inference procedure.

\section{Experiments}
\label{sec:exp}

\textbf{Datasets.}
We evaluate \ours on two datasets.
OVAD~\cite{bravo2023ovad} covers 80 classes and 117 attributes across 2,000 images from MS-COCO~\cite{patterson2016coco}, for a total of 14,300 instances. Attributes are manually annotated as \textit{positive}, \textit{negative}, and \textit{unknown}. VAW~\cite{pham2021vaw} has more concepts, \ie, 2,260 classes and 620 attributes across 10,392 images. VAW is more sparse than OVAD, \ie, around 1.8 positive and 5.3 negative attributes per instance, and it is automatically annotated.

\noindent\textbf{Setting.}
Following
prior works~\cite{bravo2023ovad, chen2023ovarnet},
\ours focuses on the box-given setting, where the bounding boxes of objects are given, as this allows us to assess its attribute detection capabilities more precisely.
However, \cref{table:box_free_ovad} also evaluates \ours in the box-free setting by introducing an object detector.
See the \textit{Supp. Mat.} for more details on this setting.

\noindent\textbf{Metrics.}
We measure performance as mean Average Precision (mAP), and also report mAP scores on \textit{head}, \textit{medium}, and \textit{tail} classes based on their frequency in the dataset.

\begin{table*}[htpb!]
\centering
\begin{tabular}{crccccccccccc}
\toprule

& \multicolumn{1}{c}{\multirow[b]{2}{*}{Method}} & \multicolumn{1}{c}{\multirow[b]{2}{0.7cm}{\centering Train\\size}} && \multicolumn{4}{c}{OVAD~\cite{bravo2023ovad}} && \multicolumn{4}{c}{VAW~\cite{pham2021vaw}} \\
\cmidrule{5-8} \cmidrule{10-13}
&&&& mAP & Head & Medium & Tail && mAP & Head & Medium & Tail \\

\midrule

\multirow{3}{*}{\rotatebox{90}{\parbox{0.8cm}{\centering Zero\\shot}}} &

CLIP~\cite{cherti2023reproducible} RN50 &&& 11.8 & 41.0 & 11.7 & 1.4 && 35.3 & 37.8 & 35.1 & 26.2 \\
& CLIP~\cite{cherti2023reproducible} ViT-B/32 &&& 17.0 & 44.3 & 18.4 & 5.5 && 50.0 & 51.0 & 50.9 & 43.2 \\
& CLIP~\cite{cherti2023reproducible} ViT-L/14 &&& 18.3 & 44.4 & 20.5 & 6.4 && 51.0 & 51.4 & 52.5 & 45.0 \\

\midrule

& \multirow{1}{*}{Image-based}
    &&& 16.8 & 45.8 & 19.2 & 3.5 && 52.9 & 53.7 & 53.8 & 46.4 \\

& TIP-Adapter~\cite{zhang2022tip} + IAP~\cite{lampert2013attribute}
        &&& 15.1 & 43.3 & 17.8 & 1.7 && 27.7 & 32.4 & 25.5 & 16.6 \\

& \multirow{1}{*}{TIP-Adapter~\cite{zhang2022tip}}
        &&& 16.7 & 44.4 & 19.7 & 3.1 && 57.5 & 57.5 & 59.4 & 51.2 \\

& \multirow{1}{*}{SuS-X~\cite{udandarao2023sus}}
        &&& 20.2 & 48.9 & 24.6 & 4.5 && 30.2 & 33.8 & 28.8 & 20.3 \\

\rowcolor{DrawioGreen} \multirow{-5}{*}{\rotatebox{90}{Cache-based}} \cellcolor{white} & \multicolumn{1}{r}{\ours}
        &&& \textbf{27.4} & \textbf{54.3} & \textbf{34.6} & \textbf{9.0} && \textbf{58.1} & \textbf{58.2} & \textbf{59.9} & \textbf{51.7} \\

\midrule

\multirow{4}{*}{\rotatebox{90}{\parbox{1.5cm}{\centering Training\\based}}}

& OVAD~\cite{bravo2023ovad} & 110k && 21.4 & 48.0 & 26.9 & 5.2 && - & - & - & - \\

& OvarNet~\cite{chen2023ovarnet} & 190k && 28.6 & 58.6 & 35.5 & 9.5 && 68.5 & - & - & - \\

& ArtVLM~\cite{zhu2024artvlm} & N/A && - & - & - & - && 71.9 & 75.0 & 72.1 & 59.4 \\

& LOWA~\cite{guo2023lowa} & 1.33M && 18.7 & 58.0 & 20.4 & 2.6 && 42.6 & 46.4 & 41.0 & 32.9 \\

\bottomrule
\end{tabular}
\caption{
    \textbf{Comparison with state of the art.}
    \inlineColorbox{DrawioGreen}{Green} indicates \ours.
    \textbf{Bold} indicates best among training-free methods.
    {The symbol ``-'' indicates results for the competitors are not available on the original papers or the impossibility to run their method due to lack of public code and/or model weights.}
See the \textit{Supp. Mat.} for an extended version of the table.
}
\label{table:additional_baselines}
\vspace{-5mm}
\end{table*}

\noindent\textbf{Baselines and competitors.}
We evaluate \ours against training-based methods, \ie, OVAD~\cite{bravo2023ovad}, OvarNet~\cite{chen2023ovarnet}, ArtVLM~\cite{zhu2024artvlm}, and LOWA~\cite{guo2023lowa}.
As we are the first to address the problem in a completely training-free fashion, we extend the baselines with several zero-shot methods~\cite{cherti2023reproducible, zhai2023sigmoid, yu2022coca, zeng2021multi, li2022blip}, and adapt previous cache-based approaches to our task~\cite{zhang2022tip, udandarao2023sus}.
For TIP-Adapter~\cite{zhang2022tip} and SuS-X~\cite{udandarao2023sus} we construct a $K$-shot cache, considering only attributes for sampling and disregarding their co-occurrences with objects. This follows their approach, which samples $K$ shots for each target \textit{class}, \ie, attributes in our task.
In addition, we test IAP~\cite{lampert2013attribute, lampert2009learning}, which conditions predicted attributes on the inferred object category, using TIP-Adapter as the object classifier.
We also consider a naive cache-based approach, \ie, ``image-based'', that uses CC12M~\cite{changpinyo2021cc12m} to build a cache specifically for each input image by retrieving the most similar samples.
Since the retrieved images are unlabeled, we use our soft labels to annotate the cache.
We construct all the proposed baselines on top of CLIP ViT-B/32~\cite{cherti2023reproducible}. 
For additional details, please refer to the \textit{Supp. Mat.}

\noindent\textbf{Implementation details.}
We use CLIP ViT-B/32~\cite{cherti2023reproducible} pre-trained on LAION2B~\cite{schuhmann2022laion} as our primary VLM.
Following TIP-Adapter~\cite{zhang2022tip}, we set $\lambda=1.17$ in \cref{eq:final-score} and $\beta=1.0$ in \cref{eq:cache-tip}.
We set $\alpha=0.6$ to mix hard and soft labels on the validation set of VAW~\cite{pham2021vaw}.
Lastly, we use CC12M~\cite{changpinyo2021cc12m} as the database in \cref{eq:dataset-stat} and GPT 3.5 Turbo~\cite{openai2022chatgpt} as the LLM for \cref{eq:llm-stat}, as we find it to perform better than GPT 4o-mini (see the \textit{Supp. Mat.}). Unless otherwise stated, we use $K=16$ shots per attribute, retrieving them from CC12M.
We use the same hyperparameters for both datasets.
For the box-free experiments, we use YOLOv11M~\cite{yolo11} as the detector.

\subsection{Results}
\label{sec:exp:results}

We present our comprehensive experimental evaluation comparing \ours with state-of-the-art training- and cache-based methods. We also show its performance with different backbones and in the box-free setting. Moreover, we discuss qualitative results and limitations of other approaches.

\noindent\textbf{Comparison with the state of the art.}
\cref{table:additional_baselines} reports the results comparing zero-shot baselines, cache-based methods, and \ours in the box-given setting.
On both benchmarks, \ours outperforms all the zero-shot and cache-based competitors, demonstrating the effectiveness of our specialized design of the cache.
It scores an increment of +10.6 mAP on OVAD and +5.2 on VAW compared to the image-based baseline.
Similarly, it surpasses TIP-Adapter~\cite{zhang2022tip} on both benchmarks by +10.7 mAP and +0.6, respectively.
SuS-X~\cite{udandarao2023sus} increases performance on OVAD while decreasing it on VAW. Overall, \ours has a gain of +7.2 mAP and +27.9 on the two datasets.
Lastly, TIP-Adapter + IAP~\cite{lampert2013attribute} demonstrates that naively using co-occurrences of attributes with objects~\cite{lampert2013attribute} is insufficient to address this task. \ours surpasses it by +12.3 mAP on OVAD and +30.4 on VAW.

We also compare \ours with training-based methods
Notably, \ours outperforms LOWA~\cite{guo2023lowa} on both benchmarks, despite the latter having been trained on a large-scale dataset while our approach requires no training data.

\noindent\textbf{Experiments with multiple VLMs.} To demonstrate that \ours is model-agnostic, we test it on various backbones and VLMs~\cite{cherti2023reproducible, yu2022coca, zhai2023sigmoid, li2022blip, zeng2021multi} and report the results in \cref{table:main_results}.
On both OVAD and VAW, \ours boosts performance consistently for all the backbones up to +11.5 mAP on OVAD with CoCa ViT-L/14~\cite{yu2022coca} and +15.8 mAP on VAW with CLIP ResNet50 (RN50)~\cite{cherti2023reproducible}. On average, \ours improves the performance of the original VLM by +8.9 on OVAD and +7.0 on VAW,
showing \ours has consistent advantages.

\begin{table*}[t!]
\parbox[t][][t]{\columnwidth}{
\centering
    \resizebox{\columnwidth}{!}{
    \begin{tabular}[t]{@{}rccccccc}
        \toprule
        \multicolumn{2}{c}{\multirow[c]{1}{*}{Method}} && \multicolumn{2}{c}{OVAD~\cite{bravo2023ovad}} && \multicolumn{2}{c}{VAW~\cite{pham2021vaw}} \\
        \cmidrule{1-2} \cmidrule{4-5} \cmidrule{7-8}
        Backbone & \ours && mAP & $\Delta$ && mAP & $\Delta$ \\
        \midrule
        &  && 11.8 & && 35.3 & \\
        \rowcolor{DrawioGreen} \cellcolor{white} \multirow[c]{-2}{*}{\raggedleft CLIP RN50~\cite{cherti2023reproducible}} & \ding{51} && 19.2 & {+7.4} && 51.1 & {+15.8} \\
        &  && 17.0 & && 50.0 & \\
        \rowcolor{DrawioGreen} \cellcolor{white} \multirow[c]{-2}{*}{\raggedleft CLIP ViT-B/32~\cite{cherti2023reproducible}} & \ding{51} && 27.4 & {+10.4} && 58.1 & {+8.1} \\
        &  && 18.3 & && 51.0 & \\
        \rowcolor{DrawioGreen} \cellcolor{white} \multirow[c]{-2}{*}{\raggedleft CLIP ViT-L/14~\cite{cherti2023reproducible}} & \ding{51} && 24.8 & {+6.5} && 61.0 & {+10.0} \\
        &  && 13.7 & && 52.5 & \\
        \rowcolor{DrawioGreen} \cellcolor{white} \multirow[c]{-2}{*}{\raggedleft SigLIP ViT-B/16~\cite{zhai2023sigmoid}} & \ding{51} && 22.1 & {+8.4} && 59.8 & {+7.3} \\
        &  && 15.1 & && 49.6 & \\
        \rowcolor{DrawioGreen} \cellcolor{white} \multirow[c]{-2}{*}{\raggedleft CoCa ViT-B/32~\cite{yu2022coca}} & \ding{51} && 24.4 & {+9.3} && 51.4 & {+1.8} \\
        &  && 14.4 & && 49.8 & \\
        \rowcolor{DrawioGreen} \cellcolor{white} \multirow[c]{-2}{*}{\raggedleft CoCa ViT-L/14~\cite{yu2022coca}} & \ding{51} && 25.9 & {+11.5} && 54.5 & {+4.7} \\
        &  && 15.9 & && 51.0 & \\
        \rowcolor{DrawioGreen} \cellcolor{white} \multirow[c]{-2}{*}{\raggedleft BLIP~\cite{li2022blip}} & \ding{51} && 22.7 & {+6.8} && 56.0 & {+5.0} \\ %
        &  && 17.4 & && 54.8 & \\
        \rowcolor{DrawioGreen} \cellcolor{white} \multirow[c]{-2}{2.3cm}{\raggedleft X-VLM~\cite{zeng2021multi}} & \ding{51} && 28.4 & {+11.0} && 57.7 & {+2.9} \\ %
        \bottomrule
    \end{tabular}
    }
    \caption{
        \textbf{Box-given results with different backbones.}
        \inlineColorbox{DrawioGreen}{Green} indicates \ours applied on top of the backbone.
        $\Delta$ indicates improvements \wrt the corresponding  baseline.
        See the \textit{Supp. Mat.} for an extended version of the table.
    }
    \label{table:main_results}
}
\hfill
\parbox[t][][t]{\columnwidth}{
\centering
    \resizebox{\columnwidth}{!}{
    \begin{tabular}[t]{@{}lrcccc}
        \toprule
        & \multicolumn{1}{c}{\multirow[b]{2}{*}{Method}} & \multicolumn{4}{c}{OVAD~\cite{bravo2023ovad}} \\
        \cmidrule{3-6}
        & & mAP & Head & Medium & Tail \\
        \midrule
        \multirow[c]{3}{*}{\rotatebox{90}{\parbox{1cm}{\centering Zero\\shot}}} &
        CLIP~\cite{cherti2023reproducible} RN50 & 12.3 & 40.2 & 12.7 & 1.9 \\
        & CLIP~\cite{cherti2023reproducible} ViT-B/32 & 12.5 & 39.6 & 12.5 & 2.7 \\
        & CLIP~\cite{cherti2023reproducible} ViT-L/14 & 13.4 & 39.7 & 14.3 & 2.9 \\
        \midrule
        & Image-based & 9.4 & 35.4 & 8.7 & 1.0 \\
        & TIP~\cite{zhang2022tip} + IAP~\cite{lampert2013attribute} & 9.8 & 36.9 & 9.2 & 1.0 \\
        & TIP-Adapter~\cite{zhang2022tip} & 10.1 & 36.0 & 9.8 & 1.2 \\
        & SuS-X~\cite{udandarao2023sus} & 15.3 & 43.0 & 17.2 & 3.2 \\
        \rowcolor{DrawioGreen} \cellcolor{white} \multirow[c]{-5}{*}{\rotatebox{90}{Cache-based}} & \ours & \textbf{21.5} & \textbf{48.4} & \textbf{26.4} & \textbf{6.0} \\
        \midrule
        \multirow[c]{3}{*}{\rotatebox{90}{\parbox{1.2cm}{\centering Training\\based}}} &
        OVAD~\cite{bravo2023ovad} & 18.8 & 47.7 & 22.0 & 4.6 \\
        & OvarNet~\cite{chen2023ovarnet} & 27.2 & 56.8 & 33.6 & 8.9 \\
        & LOWA~\cite{guo2023lowa} & 18.7 & 58.0 & 20.4 & 2.6 \\
        \bottomrule
    \end{tabular}
    }
    \caption{
        \textbf{Results in the box-free setting.}
        \inlineColorbox{DrawioGreen}{Green} indicates \ours.
        \textbf{Bold} indicates the best results among training-free methods.
        See the \textit{Supp. Mat.} for an extended version of the table.
    }
    \label{table:box_free_ovad}
}
\vspace{-2mm}
\end{table*}

\noindent\textbf{Cross-datasets advantages.}
To demonstrate the generalization capabilities of our method, we report cross-dataset results in \cref{fig:cross_dataset}.
We compare \ours with OVAD when trained on various datasets (\ie, VAW~\cite{pham2021vaw}, OVAD~\cite{bravo2023ovad}, MS-COCO~\cite{patterson2016coco}) and tested on the OVAD (left) and on the VAW (right) benchmarks.
As shown in \cref{fig:cross_dataset}, performance significantly degrades when OVAD is tested on datasets different from the training one (yellow bars vs red bar), whereas our training-free \ours achieves competitive performance (green bars) with the same RN50 backbone.
This reinforces our training-free method, which does not inherit biases from a training set, differently from training-based approaches.

\begin{figure*}[htpb!]
\centering
\includegraphics[width=\linewidth]{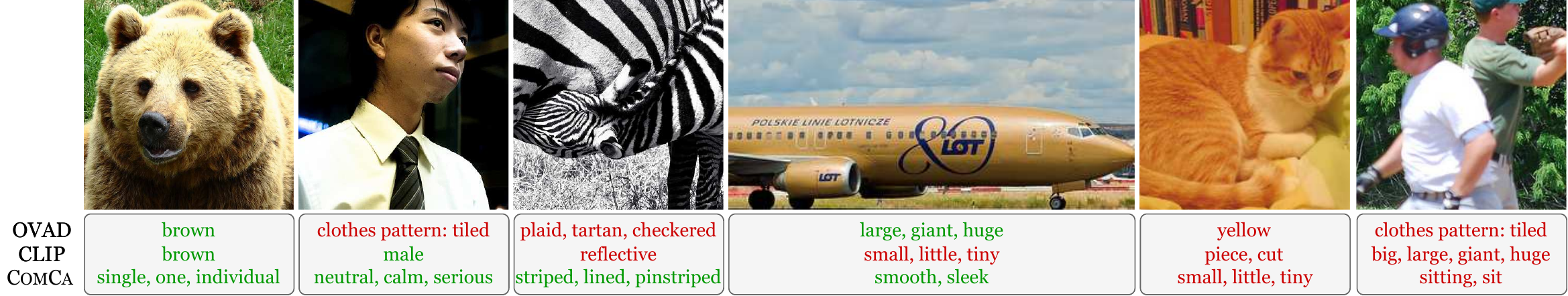}
\caption{
    \textbf{Qualitative results}.
    Predictions of OVAD, CLIP and \ours on some OVAD images.
    \inlineColorbox{DrawioGreen}{Green} are correct ones, \inlineColorbox{DrawioRed}{red} are wrong.
}
\label{fig:qualitative_results}
\vspace{-5mm}
\end{figure*}

\begin{figure}[t!]
\centering
\includegraphics[width=\columnwidth, trim={0cm 0.3cm 0cm 0cm}, clip]{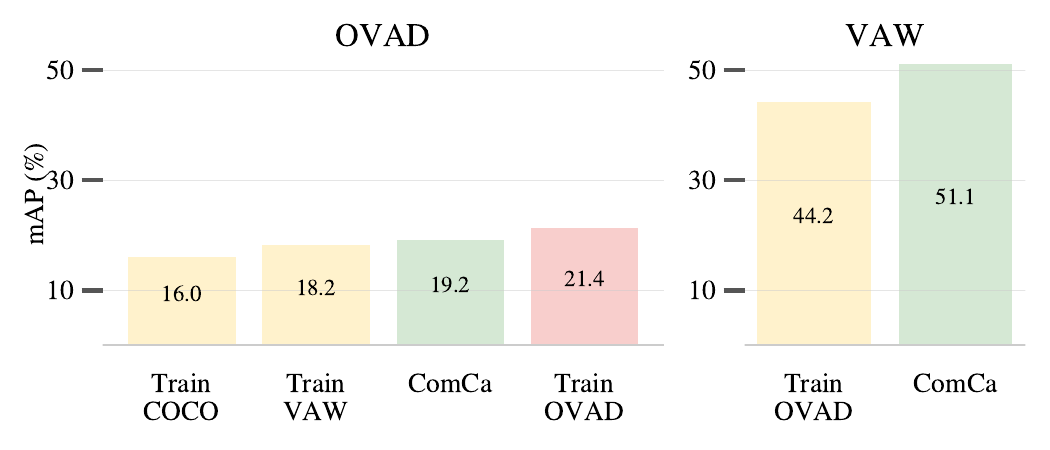}
\caption{\textbf{Cross-dataset results.}
Results on OVAD (left) and VAW (right). \inlineColorbox{DrawioYellow}{Yellow} indicates methods tested cross-domain, \ie, train and test set differ. \inlineColorbox{DrawioRed}{Red} indicates methods tested in-domain, and \inlineColorbox{DrawioGreen}{green} our training-free method, which does not suffer from domain shift problems.
All experiments use a RN50 backbone.
}
\label{fig:cross_dataset}
\vspace{-5mm}
\end{figure}

 \noindent
\textbf{Box-free setting.}
In this work, we primarily focus on attribute recognition and therefore adopt the box-given setting.
Here, we evaluate \ours in the box-free setting to demonstrate that it achieves competitive performance in a less constrained scenario.
\cref{table:box_free_ovad} shows the results for zero-shot and cache-based methods,
Specifically, we use CLIP ViT-B/32~\cite{cherti2023reproducible}
and YOLOv11M~\cite{yolo11} as detector for all models.
We note that all cache-based methods apart from SuS-X underperform the zero-shot baseline, with a decrease of up to -3.1 mAP for image-based. Only, SuS-X has a gain of +2.8 mAP \wrt CLIP ViT-B/32, with our method \ours being superior and surpassing CLIP ViT-B/32 by +9.0 mAP. Notably, our model is competitive with the training-based ones, surpassing the performance of both OVAD (+2.7 mAP) and LOWA (+2.8 mAP). 
We provide a more detailed analysis and an extended version of these results in the \textit{Supp. Mat.}

\noindent\textbf{Qualitative results.}
\cref{fig:qualitative_results}, we show qualitative examples
comparing training-based OVAD~\cite{bravo2023ovad}, CLIP ViT-B/32~\cite{cherti2023reproducible}, and \ours with the same backbone.
We consider the top-1 prediction for all methods.
OVAD correctly identifies the bear's color but misclassifies the cat's color and wrongly predicts a ``tiled'' clothes pattern for two people.
We speculate it likely predicts the same attribute for two objects (\ie, the $2^{nd}$ and the $6^{th}$) due to biases acquired during training.
CLIP makes more errors, such as labeling the zebra as ``reflective'' and the airplane as ``tiny'', while misclassifying the baseball player as ``giant''.
\ours instead understands attributes better,
for instance by accurately identifying the zebra's texture.
These issues are less pronounced in \ours thanks to our caching mechanism, which allows it to compare inputs to self-labeled examples at test time.

\subsection{Ablation study}
\label{sec:ablation}
\begin{figure*}[t!]
\parbox[t][][t]{0.335\linewidth}{
\centering
    \resizebox{\linewidth}{!}{
    \begin{tabular}[t]{lcc}
        \toprule
        \multicolumn{1}{c}{\multirow[b]{2}{*}{Configuration}} & \multicolumn{2}{c}{mAP} \\
        \cmidrule{2-3}
        & OVAD~\cite{bravo2023ovad} & VAW~\cite{pham2021vaw} \\
        \midrule
        \multicolumn{1}{l}{Zero-shot~\cite{cherti2023reproducible}} & 17.0 & 50.0 \\
        \midrule
        Random & 16.7 & \textbf{57.5} \\ %
        Brute force & 10.6 & - \\ %
        DB scores & 18.6 & 54.1 \\ %
        LLM scores & 21.1 & 55.4 \\ %
        \rowcolor{DrawioGreen} DB + LLM scores & \textbf{26.4} & 55.2 \\
        \bottomrule
    \end{tabular}
    }
    \captionof{table}{
        \textbf{Ablation on cache construction}.
        \inlineColorbox{DrawioGreen}{Green} indicates our default configuration.
        \textbf{Bold} indicates the best results.
        Results are without \textit{soft labels}.
        See the \textit{Supp. Mat.} for an extended version of the table.
    }
    \label{table:ablation_cache}
}
\hfill
\parbox[t][][t]{0.335\linewidth}{
\centering
    \resizebox{\linewidth}{!}{
    \begin{tabular}[t]{lcc}
        \toprule
        \multicolumn{1}{c}{\multirow[b]{2}{*}{Configuration}} & \multicolumn{2}{c}{mAP} \\
        \cmidrule{2-3}
        & OVAD~\cite{bravo2023ovad} & VAW~\cite{pham2021vaw} \\
        \midrule
        \multicolumn{1}{l}{Zero-shot~\cite{cherti2023reproducible}} & 17.0 & 50.0 \\
        \midrule
        {One-hot} & 26.4 & 55.2 \\
        Soft & 26.7 & 53.2 \\
        - Sharpening~\cite{assran2021semi} & 21.5 & 27.7 \\
        - Softmax & 24.5 & \textbf{58.5} \\
        \rowcolor{DrawioGreen} - \cref{eq:cache-soft} & \textbf{27.4} & 58.1 \\ %
        \bottomrule
    \end{tabular}
    }
    \captionof{table}{
        \textbf{Ablation on soft labels}.
        \inlineColorbox{DrawioGreen}{Green} indicates our default configuration. \textbf{Bold} indicates the best results.
        See the \textit{Supp. Mat.} for an extended version of the table.
    }
    \label{table:ablation_soft}
}
\hfill
\parbox[t][][t]{0.29\linewidth}{
\vspace*{0pt}
\centering
    \includegraphics[width=\linewidth, trim=0 0 0 0, clip, valign=t]{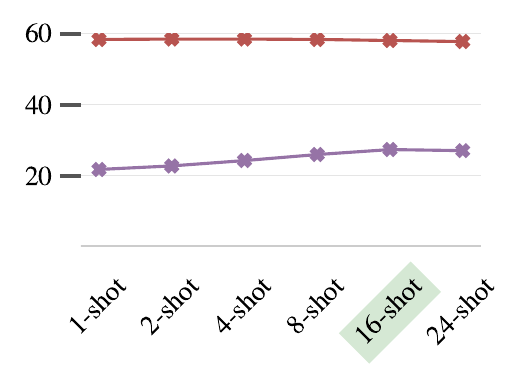}
    \captionof{figure}{
        \textbf{Ablation on number of samples per attribute.}
        The \inlineColorbox{DrawioPurple}{purple} line is mAP on OVAD, while the \inlineColorbox{DrawioRed}{red} line is mAP on VAW.
        \inlineColorbox{DrawioGreen}{Green} indicates our default configuration.
    }
    \label{fig:k-ablation}
}
\end{figure*}

\begin{figure}[t!]
    \centering
    \includegraphics[width=\columnwidth, trim=0 0 0 0, clip]{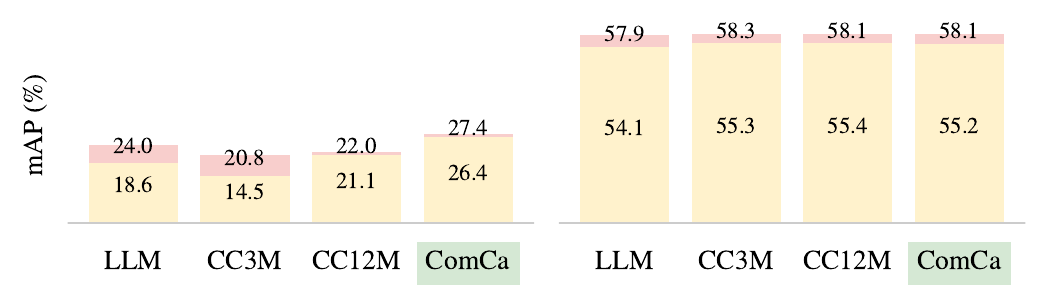}
    \caption{
        \textbf{Soft labels across cache priors.}
        Performance on OVAD (left) and VAW (right), considering generative (LLM) and retrieval (CC3M, CC12M) scores. \inlineColorbox{DrawioYellow}{Yellow} indicates one-hot labels,
        \inlineColorbox{DrawioRed}{red} soft labels and
        \inlineColorbox{DrawioGreen}{green} \ours, combining LLM and CC12M.
    }
    \label{fig:soft-vs-cache}
    \vspace{-3mm}
\end{figure}

In this section, we analyze \ours's components, studying how sampling strategies affect performance (\cref{table:ablation_cache}), soft labeling's impact (\cref{table:ablation_soft,fig:soft-vs-cache}), and the number of samples in the cache (\cref{fig:k-ablation}). Additional experiments in the \textit{Supp. Mat.} show the effect of (i) the hyperparameters $\alpha$ and $\lambda$, (ii) constructing the cache with Stable Diffusion, and (iii) using other LLMs for generative scores.

\noindent\textbf{Cache construction.}
In \cref{table:ablation_cache}, we evaluate five strategies to construct the cache.
The first, \textit{Random}, samples $K$ objects for each attribute and creates a cache of $\lvert A \rvert \times K$ examples, seeking only a match on the attribute.
The second, \textit{Brute force}, considers all possible combinations of attributes and objects, resulting in a large cache of $|O| \times |A| \times K$ samples.
As VAW contains many attributes and objects, the \textit{Brute force} cache would be too large and cannot be practically tested.
As shown in \cref{table:ablation_cache}, both approaches lead to very low results on OVAD.
Compared to the zero-shot CLIP baseline, performance is similar (\textit{Random}, -0.3 mAP) or even below (\textit{Brute force}, -6.4 mAP).
We hypothesize that \textit{Random} ignores attribute-object relationships, while \textit{Brute force} introduces many noisy and potentially detrimental samples, harming performance.
On the contrary, \textit{Random} improves by +7.5 on VAW. We hypothesize this is due to the nature of the dataset, which contains many attributes and objects.

When introducing priors on object-attributes compositions via a \textit{database} (\cref{eq:dataset-stat}) or an LLM (\cref{eq:llm-stat}), performance largely increases by up to +4.1 mAP on VAW \wrt the baseline for \textit{database scores}.
\ours combines the two scores to obtain object priors, achieving the best performance.
These results confirm both (i) that attribute-object compatibility is important to build the cache, and (ii) that databases and LLMs provide complementary information. 

\noindent\textbf{Soft labels.}
As discussed in \cref{sec:soft-caching}, each image in the cache contains multiple attributes, even if we fill the cache considering only one of them.
Here we analyze the soft labels obtained via \cref{eq:cache-soft-init}, comparing them with the \textit{One-hot} attribute labels of the initial cache, reporting the \mbox{results in \cref{table:ablation_soft}.}

The first strategy is to incorporate the soft scores as they are predicted (\textit{Soft} in \cref{table:ablation_soft}).
This leads to results slightly above the \textit{One-hot} baselines on OVAD (+0.3 mAP), while reducing the performance in VAW (-2.0 mAP).
This is a direct consequence of allowing each sample to have a different level of relevance in the cache, something that becomes more prominent in VAW due to the larger number of attributes.

To fix this issue, we can normalize the scores. 
We study three strategies: \textit{Sharpening}, where the scores are normalized as in PAWS~\cite{assran2021semi};
\textit{Softmax} normalization over each sample; and the one we propose in \cref{eq:cache-soft}, \ie, using the statistics of the cache-attributes similarity.
\textit{Sharpening} is not effective in this scenario, achieving the lowest performance across both settings.
\textit{Softmax} normalization leads to good results on VAW, \eg, +3.3 mAP \wrt \textit{One-hot}, but it struggles on OVAD (24.5 mAP) due to the limited range of values.

Our strategy addresses both issues simultaneously, consistently improving metrics in both settings compared to the \textit{One-hot} strategy, scoring 27.4 mAP on OVAD and 58.1 mAP on VAW.
Note that these findings are general. \cref{fig:soft-vs-cache} shows soft labels consistently enhance results for estimating attribute-object priors across all strategies and datasets. Improvements range from +3.8 mAP on VAW to +5.4 mAP on OVAD in the \textit{LLM}-only setting.

\noindent\textbf{Number of samples per attribute.}
In \cref{fig:k-ablation}, we measure how adding more samples (varying $K$) per attribute influences performance on the two datasets.
We observe performance consistently increasing for OVAD while going from 1 to 16 samples, reaching a plateau when moving to $K=24$.
On VAW, performance is stable for different values of $K$ due to the large number of attributes in the dataset ($620$), making the cache representative enough even when $K=1$. These results confirm the stability of \ours.

\section{Conclusion}
We introduced \oursFull for open-vocabulary attribute detection.
\ours uses a VLM and an auxiliary cache of images to solve this task in a training-free manner.
Differently from previous works~\cite{zhang2022tip, udandarao2023sus}, we construct the cache based on attribute-object compatibility derived from web-scale databases and LLMs.
We also introduced a soft-labeling scheme to account for the co-occurrence of multiple attributes within an image.
Combining these two components allows to model the natural co-occurrence of attributes in the real world, together and across objects, while making the cache construction scalable.
Our experiments demonstrate that \ours is a strong training-free approach. 
Since \ours is training-free, it does not suffer from differences between training and test attributes, contrary to training-based methods.
This also enables \ours to be adapted to special domains (\eg, aerial images) more easily than training-based methods, which would require specific re-training.
Future work will focus on devising methods to address attribute detection even in more challenging settings where objects or attributes are not given~\cite{conti2024vocabulary}.

\noindent \textbf{Acknowledgments}.
We acknowledge ISCRA for awarding this project access to the LEONARDO supercomputer, owned by the EuroHPC Joint Undertaking, hosted by CINECA (Italy).
This work was sponsored by the MUR PNRR project FAIR - Future AI Research (PE00000013), funded by NextGeneration EU, and the EU projects AI4TRUST (No.101070190) and ELIAS (No.01120237).

{
    \small
    \bibliographystyle{ieeenat_fullname}
    \bibliography{main}
}

\clearpage
\setcounter{page}{1}
\setcounter{table}{5}
\setcounter{figure}{6}
\setcounter{section}{5}
\maketitlesupplementary

In the following, we provide additional technical and implementation details, and further analyses. %
Specifically:
\begin{itemize}
    \item \cref{appendix:related_works_baselines} describes how we implemented and tested the baselines presented in Sec.~4 (\textit{Main}). %
    
    \item \cref{appendix:different-backbones} presents detailed results with different backbone VLMs.
    
    \item \cref{appendix:box_free_setting} provides a detailed description of the box-free setting, including how we adapted zero-shot and cache-based models to it, as well as extended results.

    \item \cref{appendix:additional_ablations} introduces additional ablation studies to those presented in Sec.~4.2 (\textit{Main}), and provides extended results for the ablations in \textit{Main}. %

    \item \cref{appendix:additional_qualitatives} offers additional qualitative results.

    \item \cref{appendix:prompt_details} lists all the prompt templates that \ours uses.

    \item \cref{appendix:resources-used} provides details on the hardware employed to conduct experiments.

    \item \cref{appendix:limitations} comments on the limitations of \ours, and provides possible future research directions to address them and improve attribute-detection models' capabilities.
\end{itemize}

\section{Baselines}
\label{appendix:related_works_baselines}

\begin{table*}[t!]
\centering

\resizebox{\linewidth}{!}{
\begin{tabular}{clcccccccccccc}
\toprule

& \multicolumn{1}{c}{\multirow[b]{2}{*}{\parbox{1cm}{Method}}} & \multicolumn{2}{c}{Configuration} && \multicolumn{4}{c}{OVAD~\cite{bravo2023ovad}} && \multicolumn{4}{c}{VAW~\cite{pham2021vaw}} \\
\cmidrule{3-4} \cmidrule{6-9} \cmidrule{11-14}
&& Type & Size && mAP & Head & Medium & Tail && mAP & Head & Medium & Tail \\

\midrule

\multirow{3}{*}{\rotatebox{90}{\parbox{1.2cm}{\centering Zero\\shot}}} &

CLIP~\cite{cherti2023reproducible} RN50 &&&& 11.8 & 41.0 & 11.7 & 1.4 && 35.3 & 37.8 & 35.1 & 26.2 \\
& CLIP~\cite{cherti2023reproducible} ViT-B/32 &&&& 17.0 & 44.3 & 18.4 & 5.5 && 50.0 & 51.0 & 50.9 & 43.2 \\
& CLIP~\cite{cherti2023reproducible} ViT-L/14 &&&& 18.3 & 44.4 & 20.5 & 6.4 && 51.0 & 51.4 & 52.5 & 45.0 \\

\midrule

\multirow{8}{*}{\rotatebox{90}{\parbox{4cm}{\centering Cache-based}}} &

\multirow{2}{*}{Image-based} & $\mathcal{I}$ & $K$
    && 16.8 & 45.8 & 19.2 & 3.5 && 52.9 & 53.7 & 53.8 & 46.4 \\
 && $\mathcal{I}$ & $\lvert A \rvert \times K$
    && 9.4 & 36.8 & 8.2 & 1.3 && 31.5 & 35.7 & 29.7 & 21.2 \\

\cmidrule{2-14}

& TIP-Adapter~\cite{zhang2022tip} + IAP~\cite{lampert2013attribute}
    & $\mathcal{O}$ & $\lvert O \rvert \times K$
        && 15.1 & 43.3 & 17.8 & 1.7 && 27.7 & 32.4 & 25.5 & 16.6 \\

\cmidrule{2-14}

& \multirow{3}{*}{TIP-Adapter~\cite{zhang2022tip}}
    & $\mathcal{A}$ & $\lvert A \rvert \times K$
        && 16.7 & 44.4 & 19.7 & 3.1 && 57.5 & 57.5 & 59.4 & 51.2 \\
    && $\mathcal{A}\mathcal{O}$ & $\lvert O \rvert \times \lvert A \rvert $ %
        && 9.8 & 38.3 & 8.9 & 0.9 && 53.4 & 53.8 & 54.8 & 47.7 \\
    && $\mathcal{A}\mathcal{O}$ & $\lvert O \rvert \times \lvert A \rvert \times K$
        && 10.6 & 41.2 & 9.6 & 1.0 && - & - & - & - \\

\cmidrule{2-14}

& \multirow{1}{*}{SuS-X~\cite{udandarao2023sus}}
    & $\mathcal{A}$ & $\lvert A \rvert \times K$
        && 20.2 & 48.9 & 24.6 & 4.5 && 30.2 & 33.8 & 28.8 & 20.3 \\

\cmidrule{2-14}

\rowcolor{DrawioGreen} \cellcolor{white} & \multicolumn{1}{l}{\ours}
    & $\mathcal{A}\mathcal{O}$ &  $\lvert A \rvert \times K$
        && \textbf{27.4} & \textbf{54.3} & \textbf{34.6} & \textbf{9.0} && \textbf{58.1} & \textbf{58.2} & \textbf{59.9} & \textbf{51.7} \\

\midrule

\multirow{4}{*}{\rotatebox{90}{\parbox{1.2cm}{\centering Training\\based}}} &

OVAD~\cite{bravo2023ovad} & \multicolumn{2}{c}{Train size: 110K} && 21.4 & 48.0 & 26.9 & 5.2 && - & - & - & - \\

& OvarNet~\cite{chen2023ovarnet} & \multicolumn{2}{c}{Train size: 190k} && 28.6 & 58.6 & 35.5 & 9.5 && 68.5 & - & - & - \\

& ArtVLM~\cite{zhu2024artvlm} & \multicolumn{2}{c}{Train size: N/A} && - & - & - & - && 71.9 & 75.0 & 72.1 & 59.4 \\

& LOWA~\cite{guo2023lowa} & \multicolumn{2}{c}{Train size: 1.33M} && 18.7 & 58.0 & 20.4 & 2.6 && 42.6 & 46.4 & 41.0 & 32.9 \\

\bottomrule
\end{tabular}
}

\caption{
\textbf{Comparison with state of the art.}
Extended version of Tab.~1 (\textit{Main}), reporting (i) details on \textit{head}, \textit{medium}, and \textit{tail} on OVAD and VAW, and (ii) details on the configuration used.
All cache-based models are tested with CLIP ViT-B/32~\cite{cherti2023reproducible} as backbone.
\inlineColorbox{DrawioGreen}{Green} indicates \ours.
For ArtVLM~\cite{zhu2024artvlm}, we report its best scores.
\textbf{Bold} indicates best among training-free methods.
$\mathcal{I}$ refers to images, $\mathcal{O}$ to objects, $\mathcal{A}$ to attributes, and $\mathcal{AO}$ to attribute-objects.
}
\label{supp:table:additional_baselines_full}

\end{table*}

To conduct a detailed evaluation of \ours, we implement several related works and test them in the open-vocabulary attribute detection task.
They are presented in Sec.~4 of the main paper (\textit{Main}), and this section provides additional details.
This section also presents \cref{supp:table:additional_baselines_full}, an extended version of Tab.~1 (\textit{Main}) providing (i) detailed results on all parts of the distributions of OVAD and VAW, (ii) details on the configuration used for each cache, and (iii) additional results with more configurations.

\noindent\textbf{Image-based.}
It is a simple baseline that disregards attribute-object compositions, thus making it possible to understand their effects.
In addition, it uses no textual input, as it focuses on image-to-image retrieval.
As a result, this strategy does not suffer from potential noise in prompts and/or in the text-to-image retrieval process.
The cache is constructed on a per-input basis, \ie, specific to each input image. The rationale is that, given an image, its closest ones should bear the same semantic.
Given an input image $x$, it retrieves the $K$ shots $x_c$ using no other prior knowledge, \ie, it leverages no LLM and no web-scale database to estimate attribute-object compatibility.
As a result, this method has two key characteristics:
(i) the retrieved images are the most similar to the input $x$;
(ii) it provides no labels, as image-to-image retrieving is associated to no textual prompt.
Consequently, image-based works only with our soft labeling mechanism, as it provides no one-hot labels.

\noindent\textbf{TIP-Adapter + IAP.}
This approach leverages TIP-Adapter~\cite{zhang2022tip} and IAP~\cite{lampert2009learning, lampert2013attribute} (indirect attribute prediction), using TIP-Adapter to predict the object's category, and IAP to recognize the attributes, conditioned on the predicted category.
IAP was introduced to take into account the effect of the class (\ie, object) on the attribute distribution, \ie, to condition the probability of observing an attribute given a class.
As a result, IAP exploits knowledge on the object category, thus treating attributes differently for different objects. 
IAP is defined as:
\begin{equation}
\label{eq:iap}
    p(a_m | x) = \sum_{i = 1}^{C} p(a_m | y_i) p(y_i | x),
\end{equation}
where $C$ is the number of classes and $p(y_i | x)$ is the probability of input image $x$ of belonging to class $y_i$.
We use TIP-Adapter to obtain this classification probability (\ie, $p(y_i | x)$, with the cache built only for objects), while $p(a_m | y_i)$ is the probability distribution of attribute $a_m$ of being observed together with $y_i$. This is estimated on CC12M, and it is the same that we use to construct our cache.

\noindent\textbf{TIP-Adapter.}
We directly apply TIP-Adapter~\cite{zhang2022tip}, as originally defined, to directly detect attributes.
However, since it was originally designed for object classification, we adapt it to our task by constructing its cache on attributes, rather than objects.
We populate its cache in various ways: sampling $K$ shots per attribute ($\lvert A \rvert \times K$) and sampling $K$ shots per attribute-object pair ($\lvert O \rvert \times \lvert A \rvert \times K$).
These two alternatives cover both possible ways to adapt TIP-Adapter to our task.
We note that we are unable to run the all attribute-object pairs ($\lvert O \rvert \times \lvert A \rvert \times K$) setting on VAW when $K > 1$, as memory requirements increase dramatically. We estimate some hundred gigabytes of RAM would be necessary to handle all the $2,260$ categories times $620$ attributes times $16$ shots per pair.

\noindent\textbf{SuS-X.}
Similarly, we implement SuS-X~\cite{udandarao2023sus}, a successor of TIP-Adapter that leverages a more advanced inference formulation based on the KL-divergence, while computing the scores for an input $x$.
We construct its cache by sampling $K$ images for each attribute, thus filling it with $\lvert A \rvert \times K$ samples, similarly to the original implementation.

\subsection{Full results}

In \cref{supp:table:additional_baselines_full}, we propose and compare with seven baselines on OVAD and VAW.
We test image-based by constructing a cache with (i) $K$ samples and (ii) $\lvert A \rvert \times K$ samples, with $K = 16$ in both cases.
Notably, the smaller cache performs better, scoring 16.8 mAP vs 9.4 mAP on OVAD and 52.9 mAP vs 31.5 mAP on VAW.
We hypothesize this is due to the inaccuracies in the retrieval process, which considers solely the input image $x$, with no regard to the target attributes, \eg, the object may dominate the retrieval content, producing  a cache less informative \wrt attribute labels.
The smaller cache, despite its higher similarity to the input, might contain less noisy samples, thus allowing our soft labeling scheme to produce better labels.

Next, we test IAP, using TIP-Adapter as our object classifier. The cache contains $\lvert O \rvert \times K$ elements, and we set $K = 16$. Its performance is lower than the image-based baseline's, as it achieves 15.1 mAP on OVAD (\ie, -1.7 mAP \wrt image-based) and 27.7 mAP on VAW (\ie, -25.2 mAP \wrt image-based). This is a consequence of the constrained attribute predictions, inherited from the object ones.
As $p(a_m | y_i)$ depends only on the category, different objects within the same category will have the exact same attribute probability. While the input image $x$ influences the probability $p(y_i | x)$ that it belongs to category $y_i$, it has no direct influence on $p(a_m | y_i)$. Therefore, if two objects with completely different visual attributes have the same class probabilities, this baseline will assign them the same attribute probabilities.

When including attributes in TIP-Adapter, we consider three distinct settings, covering various ways to extend its cache design to our task.
We find that the smaller cache (\ie, $\lvert A \rvert \times K$), having the same size as \ours's, is the best-performing one. However, on OVAD it underperforms its backbone when used in the zero-shot setting (16.7 vs 17.0 mAP), while it surpasses it by +7.5 mAP on VAW.

Lastly, we compare with SuS-X, which outperforms all other baselines on OVAD, achieving 20.2 mAP, \ie, +3.5 \wrt TIP-Adapter when constructing the cache in the same way.
However, on VAW it struggles, scoring only 30.2 mAP, \ie -27.3 mAP \wrt TIP-Adapter.

\ours outperforms all these baselines on both benchmarks, with a gap of up to +18.0 mAP compared to image-based on OVAD and up to +27.9 mAP \wrt SuS-X on VAW.
Specifically, on OVAD \ours surpasses TIP-Adapter by +15.0 mAP on average over all three configurations of TIP, and SuS-X by +7.2 mAP.
On VAW, \ours has a gain over image-based of +26.6 mAP and on TIP-Adapter + IAP of +30.4 mAP.
Compared to SuS-X, \ours's gap is +27.9 mAP, and obtains an average increase of +2.65 mAP \wrt TIP-Adapter, always surpassing it.

\section{Backbones}
\label{appendix:different-backbones}

\ours is a backbone-agnostic approach, therefore it supports virtually any vision-language model that can match text and images, similar to CLIP.
\cref{supp:table:backbones} is the extended version of Tab.~2, reporting detailed results on all parts of the distribution of OVAD and VAW.
Results are consistent with those of Tab.~2,  with \ours always improving the model it is applied to, across all metrics.

\begin{table*}[t!]
\centering
\resizebox{\linewidth}{!}{
    \begin{tabular}{@{}rccccccccccccc}
        \toprule

        \multicolumn{2}{c}{Method} && \multicolumn{5}{c}{OVAD~\cite{bravo2023ovad}} & \phantom{} & \multicolumn{5}{c}{VAW~\cite{pham2021vaw}} \\
        
        \cmidrule{1-2} \cmidrule{4-8} \cmidrule{10-14}
        Backbone & \ours && mAP & Head & Medium & Tail & $\Delta_{\text{mAP}}$ && mAP & Head & Medium & Tail & $\Delta_{\text{mAP}}$ \\

        \midrule

        & && 11.8 & 41.0 & 11.7 & 1.4 & && 35.3 & 37.8 & 35.1 & 26.2 \\
        \rowcolor{DrawioGreen} \cellcolor{white} \multirow[c]{-2}{*}{\raggedleft CLIP RN50~\cite{cherti2023reproducible}} & \ding{51}
        && 19.2 & 48.6 & 21.7 & 5.6 & +7.4 && 51.1 & 51.8 & 52.4 & 44.2 & +15.8 \\
        
        \addlinespace
        & && 17.0 & 44.3 & 18.4 & 5.5 & && 50.0 & 51.0 & 50.9 & 43.2 \\
        \rowcolor{DrawioGreen} \cellcolor{white} \multirow[c]{-2}{*}{\raggedleft CLIP ViT-B/32~\cite{cherti2023reproducible}} & \ding{51}
        && {27.4} & {54.3} & {34.6} & {9.0} & +10.4 && 58.1 & 58.2 & 59.9 & 51.7 & +8.1 \\
        
        \addlinespace
        & && 18.3 & 44.4 & 20.5 & 6.4 & && 51.0 & 51.4 & 52.5 & 45.0 \\
        \rowcolor{DrawioGreen} \cellcolor{white} \multirow[c]{-2}{*}{\raggedleft CLIP ViT-L/14~\cite{cherti2023reproducible}} & \ding{51}
        && 24.8 & 53.0 & 30.6 & {7.6} & +6.5 && {61.0} & {60.5} & {63.5} & {55.6} & +10.0 \\
        
        \addlinespace
        & && 13.7 & 41.5 & 15.2 & 2.0 & && 52.5 & 52.1 & 54.6 & 47.8 \\
        \rowcolor{DrawioGreen} \cellcolor{white} \multirow[c]{-2}{*}{\raggedleft SigLIP ViT-B/16~\cite{zhai2023sigmoid}} & \ding{51}
        && 22.1 & 49.5 & 27.2 & 6.1 & +8.4 && {59.8} & 59.0 & {62.4} & {54.7} & +7.3 \\
        
        \addlinespace
        & && 15.1 & 43.7 & 16.7 & 2.9 & && 49.6 & 50.7 & 50.6 & 42.6 \\
        \rowcolor{DrawioGreen} \cellcolor{white} \multirow[c]{-2}{*}{\raggedleft CoCa ViT-B/32~\cite{yu2022coca}} & \ding{51}
        && 24.4 & 50.7 & 31.7 & 6.1 & +9.3 && 51.4 & 52.1 & 52.6 & 44.7 & +1.8 \\
        
        \addlinespace
        & && 14.4 & 42.8 & 15.4 & 3.0 & && 49.8 & 50.7 & 50.9 & 42.9 \\
        \rowcolor{DrawioGreen} \cellcolor{white} \multirow[c]{-2}{*}{\raggedleft CoCa ViT-L/14~\cite{yu2022coca}} & \ding{51}
        && 25.9 & 51.7 & 33.7 & 7.1 & +11.5 && 54.5 & 54.7 & 56.2 & 48.3 & +4.7 \\
        
        \addlinespace
        & && 15.9 & 44.5 & 18.3 & 2.8 & && 51.0 & 51.3 &  52.0 & 46.6 \\
        \rowcolor{DrawioGreen} \cellcolor{white} \multirow[c]{-2}{*}{\raggedleft BLIP~\cite{li2022blip}} & \ding{51}
        && 22.7 & 51.3 & 27.3 & 7.0 & +6.8 && 56.0 & 55.8 & 58.1 & 50.6 & +5.0 \\
        
        \addlinespace
        & && 17.4 & 46.0 & 20.0 & 4.1 & && 54.8 & 56.4 & 55.7 & 46.2 \\
        \rowcolor{DrawioGreen} \cellcolor{white} \multirow[c]{-2}{*}{\raggedleft X-VLM~\cite{zeng2021multi}} & \ding{51}
        && {28.4} & {54.8} & {37.7} & 7.5 & +11.0 && 57.7 & {59.4} & 59.4 & 46.9 & +2.9 \\

        \bottomrule
    \end{tabular}
}

\caption{
\textbf{Box-given results with different backbones.}
Extended version of Tab.~2 (\textit{Main}).
\inlineColorbox{DrawioGreen}{Green} indicates \ours applied on top of the backbone.
$\Delta$ indicates improvements \wrt the corresponding baseline.
}
\label{supp:table:backbones}
\end{table*}

\section{Box-free setting}
\label{appendix:box_free_setting}

As explained in Sec.~4.1, we primarily focus on attribute recognition, thus adopting the box-given setting as our competitors~\cite{bravo2023ovad, chen2023ovarnet}.
In Tab.~3, we extend \ours to the box-free setting by introducing an object detector. %
In \cref{supp:table:box_free_ovad_full}, we show the complete version of Tab.~3, reporting the results on the \textit{head}, \textit{medium}, and \textit{tail} distributions of OVAD.
In addition, \cref{supp:table:box_free_ovad_full} reports results of all zero-shot and cache-based models with all five detectors of the YOLOv11 family~\cite{yolo11}, \ie, from the smallest ``N'' to the largest ``X''.
Specifically, we evaluate CLIP RN50~\cite{cherti2023reproducible}, CLIP ViT-B/32~\cite{cherti2023reproducible}, and CLIP ViT-L/14~\cite{cherti2023reproducible} as zero-shot models. %
For cache-based models, we report results for ``image-based'' (see Sec.~4 and \cref{appendix:related_works_baselines}), IAP~\cite{farhadi2010attribute, lampert2013attribute}, TIP-Adapter~\cite{zhang2022tip}, SuS-X~\cite{udandarao2023sus}, and \ours using CLIP ViT-B/32 as backbone.

\noindent\textbf{Implementation details.}
We introduce an object detector $f_{d}: \mathcal{X} \rightarrow \textbf{\text{box}}$, where $\textbf{\text{box}} \in \mathbb{R}^{N \times 4}$ is the set of $N$ predicted bounding boxes.
We crop the input image on each predicted bounding box,
obtaining $N$ crops, and we input each cropped image to the model, as we do in the box-given setting.

We leverage the YOLOv11 family of object detectors, testing all five models, which have an increasing number of parameters: from 2.6M for ``N'', up to 56.9M for ``X''.

\noindent\textbf{Results.}
In \cref{supp:table:box_free_ovad_full}, we evaluate \ours in the box-free setting with five different detectors and two different backbones.
Zero-shot models generally underperform their corresponding box-given counterpart, with larger models (\eg, CLIP ViT-L/14) scoring better performance than smaller models (\eg, CLIP RN50).
Interestingly, CLIP ViT-L/14 is the best-performing zero-shot model with all detectors when considering mAP, as performance ranges from 13.0 with YOLOv11N to 13.4 with YOLOv11X.
However, it is always surpassed by other models on the \textit{head} part of the distribution.
For instance, CLIP RN50 outperforms it by +0.6 on \textit{head} when using YOLOv11N.

\begin{figure*}[t!]
\centering
\includegraphics[width=\linewidth]{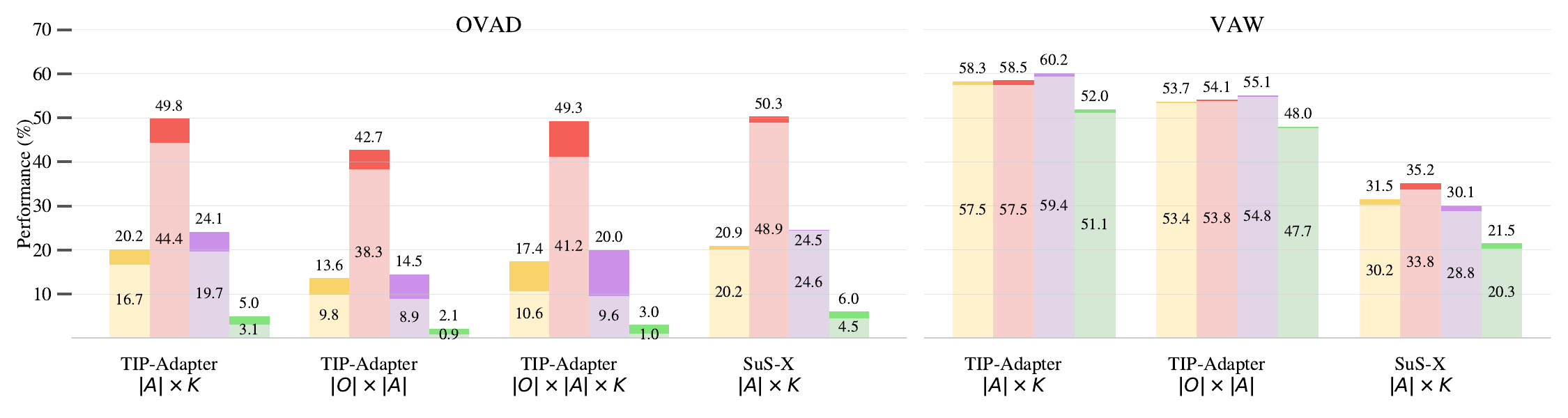}
\caption{
\textbf{Soft labels on cache-based baselines.}
We apply our soft labeling mechanism to TIP-Adapter~\cite{zhang2022tip} and SuS-X~\cite{udandarao2023sus}.
\inlineColorbox{DrawioYellow}{Yellow} represents mean average precision (mAP).
\inlineColorbox{DrawioRed}{Red} represents performance on \textit{head}, \inlineColorbox{DrawioPurple}{purple} on \textit{medium}, and \inlineColorbox{DrawioGreen}{green} on \textit{tail}.
Brighter versions of these colors represent performance with soft labels.
Number of shots $K$ is always 16 if not specified otherwise.
}
\label{fig:appendix:soft-labels-on-competitors}
\end{figure*}
\begin{table}[htpb!]
\centering

\resizebox{\columnwidth}{!}{
\begin{tabular}{@{}lcrccccc}

\toprule

& \multicolumn{1}{c}{\multirow[b]{2}{*}{YOLOv11}} &
\multicolumn{1}{c}{\multirow[b]{2}{*}{Method}} &&
\multicolumn{4}{c}{OVAD~\cite{bravo2023ovad}} \\
\cmidrule{5-8}
& & && mAP & Head & Medium & Tail \\

\midrule

\multirow{15}{*}{\rotatebox{90}{Zero-shot}} &
\cellcolor{white} & CLIP RN50~\cite{cherti2023reproducible} &&  12.0 & 39.2 & 12.5 & 1.8 \\
& \cellcolor{white} & CLIP ViT-B/32~\cite{cherti2023reproducible} && 12.4 & 38.6 & 12.8 & {2.7} \\
& \multirow[c]{-3}{*}{N} \cellcolor{white} & CLIP ViT-L/14~\cite{cherti2023reproducible} && {13.0} & 38.6 & 12.9 & {2.7} \\

\cmidrule{3-8}
& \cellcolor{white} & CLIP RN50~\cite{cherti2023reproducible} && 12.3 & 39.8 & 12.6 & 2.1  \\
& \cellcolor{white} & CLIP ViT-B/32~\cite{cherti2023reproducible} && 12.4 & 39.3 & 12.8 & 2.4 \\
& \multirow[c]{-3}{*}{S} \cellcolor{white} & CLIP ViT-L/14~\cite{cherti2023reproducible} && {13.2} & 39.3 & {14.3} & {2.5} \\

\cmidrule{3-8}
& \cellcolor{white} & CLIP RN50~\cite{cherti2023reproducible} &&  12.3 & {40.2} & 12.7 & 1.9 \\
& \cellcolor{white} & CLIP ViT-B/32~\cite{cherti2023reproducible} && 12.5 & 39.6 & 12.5 & 2.7 \\
& \multirow[c]{-3}{*}{M} \cellcolor{white} & CLIP ViT-L/14~\cite{cherti2023reproducible} && {13.4} & 39.7 & {14.3} & {2.9} \\

\cmidrule{3-8}
& \cellcolor{white} & CLIP RN50~\cite{cherti2023reproducible} &&  12.3 & {40.6} & 12.6 & 1.9 \\
& \cellcolor{white} & CLIP ViT-B/32~\cite{cherti2023reproducible} && 12.4 & 39.9 & 12.5 & 2.5 \\
& \multirow[c]{-3}{*}{L} \cellcolor{white} & CLIP ViT-L/14~\cite{cherti2023reproducible} && {13.4} & 39.9 & {14.4} & {2.7} \\

\cmidrule{3-8}
& \cellcolor{white} & CLIP RN50~\cite{cherti2023reproducible} && 12.3 & {40.7} & 12.7 & 1.8 \\
& \cellcolor{white} & CLIP ViT-B/32~\cite{cherti2023reproducible} && 12.4 & 39.9 & 12.6 & 2.4 \\
& \multirow[c]{-3}{*}{X} \cellcolor{white} & CLIP ViT-L/14~\cite{cherti2023reproducible} && {13.4} & 39.9 & {14.5} & {2.6} \\

\midrule

\multirow{25}{*}{\rotatebox{90}{Cache-based}} &
\cellcolor{white} & Image-based && 9.4 & 35.1 & 8.8 & 1.1 \\
& \cellcolor{white} & TIP-Adapter~\cite{zhang2022tip} + IAP~\cite{lampert2013attribute} && 9.9 & 36.6 & 9.4 & 1.0 \\
& \cellcolor{white} & TIP-Adapter~\cite{zhang2022tip} && 10.1 & 35.6 & 9.9 & 1.3 \\
& \cellcolor{white} & SuS-X~\cite{udandarao2023sus} && 14.9 & 41.4 & 16.7 & 3.4 \\
\rowcolor{DrawioGreen} \cellcolor{white} & \multirow[c]{-5}{*}{N} \cellcolor{white} & \ours && 20.2 & 46.3 & 24.8 & 5.4 \\

\cmidrule{3-8}
& \cellcolor{white} & Image-based && 9.4 & 35.3 & 8.8 & 1.0 \\
& \cellcolor{white} & TIP-Adapter~\cite{zhang2022tip} + IAP~\cite{lampert2013attribute} && 9.9 & 36.7 & 9.2 & 1.0 \\
& \cellcolor{white} & TIP-Adapter~\cite{zhang2022tip} && 10.1 & 35.9 & 9.8 & 1.2 \\
& \cellcolor{white} & SuS-X~\cite{udandarao2023sus} && 15.2 & 42.5 & 17.0 & 3.2 \\
\rowcolor{DrawioGreen} \cellcolor{white} & \multirow[c]{-5}{*}{S} \cellcolor{white} & \ours && 21.1 & 47.7 & 26.0 & 5.6 \\

\cmidrule{3-8}
& \cellcolor{white} & Image-based && 9.4 & 35.4 & 8.7 & 1.0 \\
& \cellcolor{white} & TIP-Adapter~\cite{zhang2022tip} + IAP~\cite{lampert2013attribute} && 9.8 & 36.9 & 9.2 & 1.0 \\
& \cellcolor{white} & TIP-Adapter~\cite{zhang2022tip} && 10.1 & 36.0 & 9.8 & 1.2 \\
& \cellcolor{white} & SuS-X~\cite{udandarao2023sus} && 15.3 & 43.0 & 17.2 & 3.2 \\
\rowcolor{DrawioGreen} \cellcolor{white} & \multirow[c]{-5}{*}{M} \cellcolor{white} & \ours && 21.5 & 48.4 & 26.4 & 6.0 \\

\cmidrule{3-8}
& \cellcolor{white} & Image-based && 9.4 & 35.6 & 8.7 & 1.0 \\
& \cellcolor{white} & TIP-Adapter~\cite{zhang2022tip} + IAP~\cite{lampert2013attribute} && 9.9 & 37.0 & 9.2 & 1.0 \\
& \cellcolor{white} & TIP-Adapter~\cite{zhang2022tip} && 10.1 & 36.1 & 9.8 & 1.2 \\
& \cellcolor{white} & SuS-X~\cite{udandarao2023sus} && 15.3 & 43.2 & 17.2 & 3.2 \\
\rowcolor{DrawioGreen} \cellcolor{white} & \multirow[c]{-5}{*}{L} \cellcolor{white} & \ours && 21.6 & 48.7 & 26.5 & 6.0 \\

\cmidrule{3-8}
& \cellcolor{white} & Image-based && 9.4 & 35.5 & 8.7 & 1.0 \\
& \cellcolor{white} & TIP-Adapter~\cite{zhang2022tip} + IAP~\cite{lampert2013attribute} && 9.9 & 37.1 & 9.2 & 1.0 \\
& \cellcolor{white} & TIP-Adapter~\cite{zhang2022tip} && 10.1 & 36.2 & 9.7 & 1.2 \\
& \cellcolor{white} & SuS-X~\cite{udandarao2023sus} && 15.4 & 43.5 & 17.2 & 3.1 \\
\rowcolor{DrawioGreen} \cellcolor{white} & \multirow[c]{-5}{*}{X} \cellcolor{white} & \ours && 21.7 & 49.0 & 26.6 & 6.1 \\

\bottomrule

\end{tabular}
}

\caption{
\textbf{Results in the box-free setting.}
Extended version of Tab.~3 (\textit{Main}).
\inlineColorbox{DrawioGreen}{Green} indicates \ours.
Experiments for cache-based models are conducted using CLIP ViT-B/32~\cite{cherti2023reproducible} as backbone.
}

\label{supp:table:box_free_ovad_full}
\end{table}

When considering cache-based models, \ours is the best-performing approach on all metrics, regardless of the backbone of choice.
\ours outperforms SuS-X by +6.0 mAP on average, across all object detectors.
When compared with the other competitors, the gap further increases: on average, +11.8 mAP \wrt image-based, +11.3 mAP \wrt to TIP-Adapter + IAP, and +11.1 mAP \wrt TIP-Adapter.

When introducing \ours, we observe a constant increase in performance as we increase the detector's size.

\section{Additional ablations}
\label{appendix:additional_ablations}

In this section, we provide the extended versions of the ablation experiments presented in Sec.~4.2 (from \cref{appendix:cache_construction,appendix:soft_labels,appendix:number_of_shots}), and we perform additional ablation studies to further evaluate the performance of \ours.
Extended results on cache construction are provided in \cref{appendix:cache_construction}, complementing the main findings in Tab.~4.
\cref{appendix:soft_labels} shows detailed results on all parts of the distribution of OVAD and VAW for the ablation on soft labels.
In \cref{appendix:number_of_shots}, we present results using varying numbers of samples per attribute, specifically more and less than 16, the value used in the main paper.
\cref{appendix:prior_combination} presents detailed results for the ablation on Eq.~(7).
Next, \cref{appendix:llms} shows how different LLMs influence performance when they provide \textit{generative scores} to estimate attribute-object distributions, as described in Sec.~3.2.
Additionally, \cref{appendix:lambda} replicates the ablation study conducted in TIP-Adapter~\cite{zhang2022tip} on the hyperparameter $\lambda$, thus studying if its effect is similar on \ours's cache.
In \cref{appendix:cache_gen_vs_ret}, we compare image retrieval (our default) with image generation.
Lastly, \cref{appendix:alpha} and \cref{appendix:norm} explore two key components of our algorithm: the hyperparameter $\alpha$, which blends soft and hard labels, and the normalization of final predictions.
Importantly, we highlight that $\alpha$ is set on VAW's validation set, while $\lambda = 1.17$ is the default value, taken from TIP-Adapter.

\subsection{Cache construction}
\label{appendix:cache_construction}

In \cref{supp:table:ablation_cache_full}, we report the complete results of the experiments presented in Tab.~4 (\textit{Main}).
We provide details on \textit{head}, \textit{medium}, and \textit{tail} parts of the distributions of the OVAD and VAW benchmarks.
We note that \textit{Brute force} cannot be run on VAW due to the size of the target space: the cache would be too large, and it cannot be practically tested.
Results are consistent with those of Tab.~4, showing the importance of attribute-object compatibility to build the cache, and with databases and LLMs providing complementary information.

We highlight that the cache construction process is fairly lightweight: image retrieval and soft scoring (\cref{appendix:soft_labels}) take $\approx 0.8$ seconds per attribute and LLM-based scoring (\cref{appendix:llms}) takes up to $10$ seconds per attribute.
Both of these operations occur \emph{before} inference and take much less time compared to training-based methods (\eg, 7 hours for ArtVLM~\cite{zhu2024artvlm}).
During inference, using the cache adds only two matrix multiplications and one scaling operation (Eq.~(10)), resulting in a negligible $+0.3\%$ time overhead.

\begin{table*}[t!]
\centering
    \begin{tabular}{lccccccccc}
        \toprule

        \multicolumn{1}{c}{\multirow[b]{2}{*}{\parbox{3cm}{\centering Configuration}}} & \multicolumn{4}{c}{OVAD~\cite{bravo2023ovad}} && \multicolumn{4}{c}{VAW~\cite{pham2021vaw}} \\
        
        \cmidrule{2-5} \cmidrule{7-10}
        & mAP & Head & Medium & Tail && mAP & Head & Medium & Tail \\

        \midrule

        \multicolumn{1}{l}{Zero-shot~\cite{cherti2023reproducible}} & 17.0 & 44.3 & 18.4 & 5.5 && 50.0 & 51.0 & 50.9 & 43.2 \\

        \midrule

        Random & 16.7 & 44.4 & 19.7 & 3.1 && \textbf{57.5} & \textbf{57.5} & \textbf{59.4} & \textbf{51.2} \\
        Brute force & 10.6 & 41.2 & 9.6 & 1.0 && - & - & - & - \\
        DB scores & 18.6 & 46.8 & 22.5 & 3.7 && 54.1 & 53.7 & 55.8 & 50.1 \\ %
        LLM scores & 21.1 & 51.5 & 25.0 & 5.6 && 55.4 & 56.0 & 57.0 & 47.9 \\ %
        \rowcolor{DrawioGreen} DB + LLM scores & \textbf{26.4} & \textbf{52.5} & \textbf{33.7} & \textbf{8.3} && 55.2 & 55.5 & 56.8 & 48.9 \\

        \bottomrule
    \end{tabular}

\caption{
\textbf{Ablation on cache construction.}
Extended version of Tab.~4 (\textit{Main}).
\inlineColorbox{DrawioGreen}{Green} indicates our configuration.
\textbf{Bold} indicates the best for each column.
Results are without \textit{soft labels}.
}

\label{supp:table:ablation_cache_full}
\end{table*}

\begin{table*}[t!]
\centering
    \begin{tabular}{clccccccccc}
        \toprule
        
        \multicolumn{2}{c}{\multirow[b]{2}{*}{\parbox{3cm}{\centering Configuration}}} & \multicolumn{4}{c}{OVAD~\cite{bravo2023ovad}} & \multicolumn{4}{c}{VAW~\cite{pham2021vaw}} \\
    
        \cmidrule{3-6} \cmidrule{8-11}
        & & mAP & Head & Medium & Tail && mAP & Head & Medium & Tail \\
    
        \midrule
    
        \multicolumn{2}{l}{Zero-shot~\cite{cherti2023reproducible}} & 17.0 & 44.3 & 18.4 & 5.5 && 50.0 & 51.0 & 50.9 & 43.2 \\
    
        \midrule
        
        & {One-hot} & 26.4 & 52.5 & 33.7 & 8.3 && 55.2 & 55.5 & 56.8 & 48.9 \\
        & Soft & 26.7 & 53.6 & 33.8 & 8.5 && 53.2 & 54.3 & 53.9 & 46.6 \\
        & - Sharpening~\cite{assran2021semi} & 21.5 & 49.4 & 26.0 & 6.0 && 27.7 & 32.4 & 25.5 & 16.6 \\
        & - Softmax
        & 24.5 & 52.2 & 30.3 & 7.6 && \textbf{58.5} & \textbf{58.7} & \textbf{60.3} & \textbf{52.4} \\
        \rowcolor{DrawioGreen} & - Eq.~(9) & \textbf{27.4} & \textbf{54.3} & \textbf{34.6} & \textbf{9.0} && 58.1 & 58.2 & 59.9 & 51.7 \\
    
        \bottomrule
    \end{tabular}

\caption{
\textbf{Ablation on soft labels.}
Extended version of Tab.~5 (\textit{Main}).
\inlineColorbox{DrawioGreen}{Green} indicates our default configuration.
\textbf{Bold} indicates the best for each column.
}
\label{supp:table:ablation_soft_full}
\end{table*}

\begin{table*}[t!]
\centering
    \begin{tabular}{clccccccccc}
        \toprule

        \multicolumn{2}{c}{\multirow[b]{2}{*}{Configuration}} & \multicolumn{4}{c}{OVAD~\cite{bravo2023ovad}} && \multicolumn{4}{c}{VAW~\cite{pham2021vaw}} \\
        
        \cmidrule{3-6} \cmidrule{8-11}
        & & mAP & Head & Medium & Tail && mAP & Head & Medium & Tail \\

        \midrule

        \multicolumn{2}{l}{Baseline zero-shot~\cite{cherti2023reproducible}}   & 17.0 & 44.3 & 18.4 & 5.5 && 50.0 & 51.0 & 50.9 & 43.2 \\

        \midrule
        
        & $K = 1$  & 21.8 & 49.2 & 26.7 & 6.1 && \textbf{58.4} & \textbf{58.6} & 60.2 & 52.2 \\
        & $K = 2$  & 22.8 & 50.7 & 27.8 & 6.8 && \textbf{58.4} & \textbf{58.6} & \textbf{60.3} & 51.8 \\
        & $K = 4$  & 24.3 & 51.9 & 29.9 & 7.7 && \textbf{58.4} & 58.5 & \textbf{60.3} & \textbf{52.1} \\
        & $K = 8$  & 26.0 & \textbf{54.6} & 32.2 & 8.2 && \textbf{58.4} & \textbf{58.6} & 60.2 & 51.9 \\
        \rowcolor{DrawioGreen} & $K = 16$ & \textbf{27.4} & 54.3 & \textbf{34.6} & \textbf{9.0} && 58.1 & 58.2 & 59.9 & 51.7 \\
        & $K = 24$ & 27.1 & \textbf{54.6} & \textbf{34.6} & 8.2 && 57.8 & 58.1 & 59.4 & 51.6 \\
        
        \bottomrule
    \end{tabular}

\caption{
\textbf{Ablation on the number of shots.}
Extended version of Fig.~5 (\textit{Main}).
\inlineColorbox{DrawioGreen}{Green} indicates our default configuration.
\textbf{Bold} indicates the best for each column.
}
\label{supp:table:ablation_num_shots}

\end{table*}

\begin{table*}[t!]
\centering
    \begin{tabular}{ccccccccccc}
        \toprule
        \multicolumn{1}{c}{\multirow[b]{2}{*}{\parbox{2cm}{\centering Configuration}}} && \multicolumn{4}{c}{OVAD~\cite{bravo2023ovad}} && \multicolumn{4}{c}{VAW~\cite{pham2021vaw}} \\
        \cmidrule{3-6} \cmidrule{8-11}
        && mAP & Head & Medium & Tail && mAP & Head & Medium & Tail \\
        
        \midrule
        
        \multicolumn{1}{l}{None}
            && 24.0 & 52.5 & 29.6 & 7.1 && 57.9 & 57.8 & 59.9 & 52.5 \\
        
        \multicolumn{1}{l}{Sum}
            && 24.8 & 52.5 & 31.8 & 6.4 && \textbf{58.3} & \textbf{58.4} & \textbf{60.1} & \textbf{52.0} \\
        
        \rowcolor{DrawioGreen}\multicolumn{1}{l}{Multiplication~--~Eq.~(7)}
            && \textbf{27.4} & \textbf{54.3} & \textbf{34.6} & \textbf{9.0} && 58.1 & 58.2 & 59.9 & 51.7 \\
        
        \bottomrule
    \end{tabular}

\caption{
\textbf{Ablation on Eq.~(7).}
Results of \ours on OVAD using no CC12M prior (LLM only), CC12M combined with the LLM scores using sum and \inlineColorbox{DrawioGreen}{multiplication}, as in Eq.~(7) (our default configuration).
\textbf{Bold} indicates the best results.
}
\label{supp:table:equation_7_full}
\vspace{15pt}
\end{table*}

\subsection{Soft labels}
\label{appendix:soft_labels}

\begin{table*}[t!]
\parbox[t][][t]{\columnwidth}{
    \resizebox{\columnwidth}{!}{
        \begin{tabular}[t]{lcccccc}
        \toprule
        
        \multicolumn{2}{c}{Prior} && \multicolumn{4}{c}{OVAD~\cite{bravo2023ovad}} \\
        \cmidrule{1-2} \cmidrule{4-7}
        LLM & CC12M && mAP & Head & Medium & Tail \\
        
        \midrule
        
        &&& 21.0 & 52.2 & 24.8 & 5.3 \\
        \multirow{-2}{*}{Gemma 7b} & \ding{51} && 22.8 & 54.7 & 27.0 & 6.3 \\

        \rowcolor{lightgray!40} &&& 21.4 & 50.4 & 25.9 & 5.6 \\
        \rowcolor{lightgray!40} \multirow{-2}{*}{LLaMa 2 7b} & \ding{51} && 23.2 & 52.4 & 28.2 & 6.8 \\
        
        &&& 22.8 & 53.8 & 27.3 & 6.2 \\
        \multirow{-2}{*}{LLaMa 3 8b} & \ding{51} && 24.1 & 54.8 & 29.0 & 7.1 \\
        
        \rowcolor{lightgray!40} &&& 23.9 & 52.8 & 29.4 & 7.0 \\
        \rowcolor{lightgray!40}\multirow{-2}{*}{Mistral} & \ding{51} && 24.6 & 53.2 & 30.5 & 7.2 \\

        \midrule

        &&& 24.0 & 52.5 & 29.6 & 7.1 \\  %
        \rowcolor{DrawioGreen}\cellcolor{white}\multirow{-2}{*}{GPT 3.5 Turbo} & \ding{51} && \textbf{27.4} & \textbf{54.3} & \textbf{34.6} & 9.0 \\

        \rowcolor{lightgray!40} &&& 23.6 & 50.9 & 28.5 & 7.9 \\  %
        \rowcolor{lightgray!40}\multirow{-2}{*}{GPT 4o-mini} & \ding{51} && 25.4 & 53.0 & 30.3 & \textbf{9.7} \\  %

        \bottomrule
        \end{tabular}
    }
    \caption{
    \textbf{Ablation on LLM backbones.}
    Comparison of different LLMs to obtain the attribute-object compatibility score.
    \ours is tested by using LLM scores only, and combining them with the prior extracted from CC12M~\cite{changpinyo2021cc12m}.
    \inlineColorbox{DrawioGreen}{Green} indicates our default configuration.
    \textbf{Bold} indicates the best performance.
    }
    \label{supp:table:ablation_llms}
}
\hfill
\parbox[t][][t]{\columnwidth}{
\centering
\rowcolors{2}{}{lightgray!40}
    \begin{tabular}[t]{lcccc}
    \toprule
    \multicolumn{1}{c}{\multirow[b]{2}{*}{$\lambda$}} & \multicolumn{4}{c}{OVAD~\cite{bravo2023ovad}} \\
    
    \cmidrule{2-5}
    & mAP & Head & Medium & Tail \\
    
    \midrule
    
    CLIP only & 17.0 & 44.3 & 18.4 & 5.5 \\
    
    \midrule
    
    0 & 20.6 & 48.7 & 25.1 & 5.3  \\
    0.5 & 25.7 & 53.0 & 32.0 & 8.2 \\
    1.0 & 27.1 & 54.1 & 34.1 & \textbf{9.0} \\
    \rowcolor{DrawioGreen} 1.17 (TIP default) & \textbf{27.4} & \textbf{54.3} & 34.6 & \textbf{9.0} \\
    2.0 & \textbf{27.4} & \textbf{54.3} & \textbf{34.9} & 8.7 \\
    3.0 & 27.2 & 54.1 & 34.8 & 8.4 \\
    4.0 & 27.1 & 53.9 & 34.7 & 8.2 \\
    
    \midrule
    
    Cache only & 25.9 & 52.8 & 33.6 & 6.9 \\
    
    \bottomrule
    \end{tabular}
    \caption{
    \textbf{Ablation on hyperparameter $\lambda$.}
    We replicate TIP-Adapter's~\cite{zhang2022tip} ablation on $\lambda$ to determine if \ours's cache behaves similarly.
    \inlineColorbox{DrawioGreen}{Green} indicates the default configuration for \ours, the same used in TIP-Adapter.
    \textbf{Bold} indicates the best performance.
    }
    \label{supp:table:ablation_lambda_hyperparameter}
}
\end{table*}

\cref{supp:table:ablation_soft_full} shows the complete results of the ablation on soft labels, presented in Tab.~5 (\textit{Main}).
The table includes scores on all parts of the distributions of OVAD and VAW.
Results are consistent with those of Tab.~5, showing that soft labels always increase performance by a significant margin.

In addition, we evaluate and demonstrate the effectiveness of our soft-labeling mechanism by applying it to the TIP-Adapter and SuS-X baselines in \cref{fig:appendix:soft-labels-on-competitors}.
We evaluate the same four settings as in \cref{supp:table:additional_baselines_full}.
Due to the impossibility of running TIP-Adapter in the $\lvert O \rvert \times \lvert A \rvert \times K$ setting, with $K = 16$, as explained in \cref{appendix:related_works_baselines}, we omit the plot in the figure.

We note that our soft labels \emph{always} improve mAP performance, with an average of +3.7 on OVAD and +0.8 on VAW.
Notably, it often leads to large performance increases on the \textit{medium} part of the distribution (up to +10.4), and it provides significant boosts on the \textit{tail} (up to +2.0).

\subsection{Number of shots}
\label{appendix:number_of_shots}

We evaluate our model's performance with different numbers of samples per attribute in the cache.
\cref{supp:table:ablation_num_shots} provides an extended version of the results shown in Fig.~5, illustrating that more samples generally improve performance.
On OVAD~\cite{bravo2023ovad}, performance increases steadily from $K=1$ to $K=16$, with a slight loss at $K=24$.
In contrast, VAW~\cite{pham2021vaw} is less sensitive to the number of samples. We hypothesize that this is due to our soft labeling mechanism, which may yield diminishing returns as the number of attributes increases (VAW has 620 attributes).
Results are consistent with those shown in Fig.~5 and confirm the ablation on the number of shots shown in TIP-Adapter~\cite{zhang2022tip}, where a modest cache of 16 elements contains all the necessary information.

\subsection{Prior combination}
\label{appendix:prior_combination}

In \cref{supp:table:equation_7_full}, we assess changes in performance when changing the interaction between the two priors~--~database and LLM.
We study three case, \textit{None}, \textit{Sum}, and \textit{Multiplication}, conducting experiments on both OVAD and VAW.
In \textit{None}, we discard the database prior, \ie, the compatibility scores extracted from CC12M.
In \textit{Sum}, we sum the scores produced by the LLM to the scores extracted from the web-scale database, while in \textit{Multiplication}, we follow our definition as in Eq.~(7) (refer to \textit{Main}).

\textit{None} is the worst strategy, because it excludes the information from the web-scale database. This is supported by the results presented in Fig.~6 in Sec.~4.1 in the \textit{Main} paper.
When introducing the retrieval scores with \textit{Sum}, we observe an increase in performance: +0.8 mAP on OVAD and +0.4 mAP on VAW.
\textit{Multiplication} largely outperforms \textit{Sum} on OVAD, surpassing it by +2.6 mAP (+3.4 mAP \wrt \textit{None}), while it performs slightly worse on VAW, scoring 58.1 mAP (-0.2 \wrt \textit{Sum}).
However, it is superior to \textit{None}, with a gap of +0.2 mAP.

\begin{figure*}[t!]
\parbox[t][][t]{0.48\linewidth}{
\centering
    \resizebox{\linewidth}{!}{
        \begin{tabular}[t]{lcccc}
            \toprule
            \multicolumn{1}{c}{\multirow[b]{2}{*}{Configuration}} & \multicolumn{4}{c}{OVAD~\cite{bravo2023ovad}} \\
            \cmidrule{2-5}
            & mAP & Head & Medium & Tail \\
            \midrule
            \rowcolor{DrawioGreen} Retrieval & & & & \\
            \rowcolor{DrawioGreen} (CC12M) & \multirow[c]{-2}{*}{27.4} & \multirow[c]{-2}{*}{54.3} & \multirow[c]{-2}{*}{34.6} & \multirow[c]{-2}{*}{9.0} \\
            Generation & \multirow[c]{2}{*}{\textbf{28.0}} & \multirow[c]{2}{*}{\textbf{57.2}} & \multirow[c]{2}{*}{\textbf{34.9}} & \multirow[c]{2}{*}{\textbf{9.1}} \\
            (Stable Diffusion XL) & \\
            \midrule
            \multicolumn{1}{r}{$\Delta$} & +0.6 & +2.9 & +0.3 & +0.1 \\
            \bottomrule
        \end{tabular}
    }
    \captionof{table}{
        \textbf{Ablation on cache construction.}
        Results on OVAD when constructing the cache with retrieval or generation.
        \inlineColorbox{DrawioGreen}{Green} indicates our default configuration, which uses retrieval to populate the cache.
        \textbf{Bold} indicates the best performance.
        $\Delta$ indicates the difference between generation and retrieval.
    }
    \label{supp:table:ablation_gen_vs_ret}
}
\hfill
\parbox[t][][t]{0.48\linewidth}{
\vspace*{0pt}
\centering
    \includegraphics[width=\linewidth]{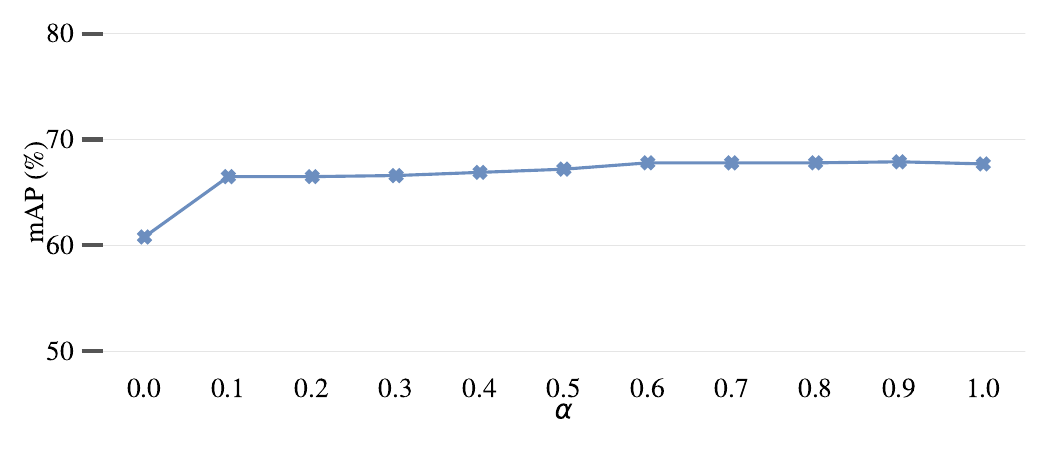}
    \captionof{figure}{
        \textbf{Ablation on hyperparameter $\alpha$.}
        The \inlineColorbox{DrawioBlue}{blue} line is mAP on the validation set of VAW.
        The higher $\alpha$, the larger the contribution of soft labels.
    }
    \label{supp:fig:plot-alpha}
}
\end{figure*}

\subsection{Different LLMs}
\label{appendix:llms}

All experiments in the main paper, along with ablations in both the \textit{Main} and \textit{Supp. Mat.}, use GPT 3.5 Turbo as the LLM for generative scores (Eq.~(6)).
In \cref{supp:table:ablation_llms}, we replace GPT 3.5 Turbo both with GPT 4o-mini and with publicly available LLMs that can be run locally.
For computational efficiency, we select models in the 7-8 billion parameters range: Gemma 7b, LLaMa 2 7b, LLaMa 3 8b, and Mistral.
We evaluate each LLM's performance using its scores alone and in combination with retrieval scores (Eq.~(5)), as described in Eq.~(7).
GPT 4o-mini underperforms GPT 3.5 Turbo, which surpasses it by +2.0 mAP.
However, GPT 4o-mini outperforms all four open-source small models, with an average gap of +1.7 mAP.
Although open source models are surpassed by the GPT family of closed models, they represent a valid open-source and cost-effective alternative.
In all cases, combining both sources yields better results, with mAP gains of $+1.8$ mAP for Gemma 7b, $+1.8$ mAP for LLaMa 2 7b, $+1.3$ mAP for LLaMa 3 8b, $+0.7$ mAP for Mistral, and +1.8 mAP for GPT 4o-mini, thus reinforcing the findings presented in Fig.~6 (\textit{Main}).

\subsection{Combining CLIP and cache scores}
\label{appendix:lambda}

By default, we use the same value of $\lambda = 1.17$ as in TIP-Adapter~\cite{zhang2022tip}, but we analyze the effect of $\lambda$ in Eq.~(11), where it is used to linearly combine vanilla CLIP scores with cache scores.
We evaluate its impact on \ours in \cref{supp:table:ablation_lambda_hyperparameter}.
The first row, \textit{CLIP only}, reports the results for the vanilla CLIP model, as seen in our baseline in Tabs.~1 and~2 (\textit{CLIP ViT-B/32}).
In the subsequent rows, increasing $\lambda$ from $0$ to $4$ shows performance gains up to $\lambda = 2.0$, with a slight decline decline afterward.
Finally, \textit{Cache only} represents using only cache scores, which, while lower than the combined approach, still outperforms \textit{CLIP only} by $+8.9$ mAP.

\subsection{Image retrieval vs generation}
\label{appendix:cache_gen_vs_ret}

Our approach involves populating the cache with images from CC12M~\cite{changpinyo2021cc12m}, selected based on cosine similarity to a text query describing an object and an attribute.
Alternatively, one could use a generative model, such as Stable Diffusion, to create samples from the same query.
To compare these methods, we conducted an ablation study contrasting web-scale image-to-text retrieval with text-to-image generation using diffusion models.
Results, shown in \cref{supp:table:ablation_gen_vs_ret}, reveal no significant performance difference, with a 0.6 mAP gap on OVAD favoring generation. Although retrieval is slightly less effective, it is far more computationally efficient.

\subsection{Alpha blending}
\label{appendix:alpha}

We evaluate the effect of the $\alpha$ parameter, which controls the blending of one-hot and soft labels, as outlined at the end of Sec.~3.2.
We set $\alpha$ to $0.6$ on the validation set of VAW, thus assigning a weight of $0.6$ to soft labels and $1 - 0.6 = 0.4$ to hard labels.

To understand the effect of this parameter, we perform experiments on the validation set of VAW~\cite{pham2021vaw}.
As shown in \cref{supp:fig:plot-alpha}, incorporating soft labels improves performance.
Starting at $\alpha = 0$, which uses only one-hot labels, we see a rapid increase in performance, peaking at $67.8$ mAP when $\alpha = 0.6$.
Beyond this point, performance remains constant, slightly decreasing to $67.7$ mAP with $\alpha = 1.0$.

We hypothesize this behavior is due to the nature of one-hot and soft labels.
One-hot labels focus solely on the attribute in the prompt, ignoring all other attributes in the image.
Relying only on hard labels reduces the task to a multiclass problem, where only one attribute is considered positive and the rest negative.
Conversely, using only soft labels yields results similar to CLIP baselines, as both approaches would use the same scoring mechanism.
The best performance is achieved by balancing the contributions of both hard and soft labels.

\begin{table*}[htpb!]
\centering

\begin{tabular}{clcccccccccc}
\toprule
\multicolumn{2}{c}{\multirow[b]{2}{*}{Configuration}} && \multicolumn{4}{c}{OVAD~\cite{bravo2023ovad}} && \multicolumn{4}{c}{VAW~\cite{pham2021vaw}} \\
\cmidrule{4-7} \cmidrule{9-12}
&&& mAP & Head & Medium & Tail && mAP & Head & Medium & Tail \\

\midrule

\multicolumn{2}{l}{Baseline zero-shot~\cite{cherti2023reproducible}}
&& 17.0 & 44.3 & 18.4 & 5.5 && 50.0 & 51.0 & 50.9 & 43.2 \\

\midrule

& $\texttt{no norm}$ && 8.7 & 36.0 & 7.4 & 0.6 && 40.3 & 43.1 & 39.6 & 31.7 \\
& $\texttt{min-max norm}$ && 25.9 & 53.4 & 32.1 & 8.67 && 56.6 & 56.9 & 58.5 & 49.7 \\
\rowcolor{DrawioGreen} & $\texttt{max norm}$, $\texttt{softmax}$ && 27.4 & 54.3 & 34.6 & 9.0 && 58.1 & 58.2 & 59.9 & 51.7 \\

\bottomrule
\end{tabular}

\caption{
\textbf{Ablation on scores normalization}.
Results with different cache score normalization strategies.
\inlineColorbox{DrawioGreen}{Green} indicates the default configuration for \ours.
}
\label{supp:tab:ablation_norm}

\end{table*}

\subsection{Cache scores normalization}
\label{appendix:norm}

In \cref{supp:tab:ablation_norm}, we analyze the effect of scaling the scores from cache-based predictions by comparing three strategies:
no normalization (\texttt{no norm}, as in~\cite{zhang2022tip}),
\texttt{min-max} normalization (subtracting the minimum score and dividing by the maximum),
and our approach of scaling by the maximum value (\texttt{max}) used in Eq.~(11).
Additionally, we apply softmax within $\eta_A$ to ensure numerical stability and produce a proper probability distribution.
The results, shown at the bottom of~\cref{supp:tab:ablation_norm}, indicate that no normalization yields poor performance ($8.7$ mAP on OVAD, $40.3$ mAP on VAW), while \texttt{min-max} improves results ($25.9$ mAP on OVAD, $56.6$ mAP on VAW).
Our method achieves the best performance overall, with $27.4$ mAP on VAW and $58.1$ mAP~on~VAW.

\subsection{Seen vs unseen bias}
\label{appendix:seen-unseen-bias}

We note that by omitting the training stage
(i) \ours is not exposed to training biases and
(ii) it is impractical to identify the seen/unseen split, as everything is technically unseen.
We use VAW's unseen split from~\cite{chen2023ovarnet} to directly compare with OvarNet.
\ours shows more consistent results (58.4 mAP on seen, 56.9 on unseen) than OvarNet (69.8 mAP on seen, 56.4 on unseen).
As potential biases may come from the cache, we check the performance on the object-attribute compositions there stored, considering as ``seen'' those present in the cache.
\ours scores 56.5 mAP (seen) and 58.2 (unseen) on VAW, and 44.8 mAP (seen) and 27.9 (unseen) on OVAD.
Note that the difference in performance between the two sets in OVAD follows the zero-shot performance of the base model (\eg, 37.8 and 18.8 mAP) rather than cache-specific biases.

\section{Additional qualitative results}
\label{appendix:additional_qualitatives}

\begin{figure*}[t!]
    \centering
    \begin{subfigure}[t]{\linewidth}
        \centering
        \includegraphics[width=\linewidth]{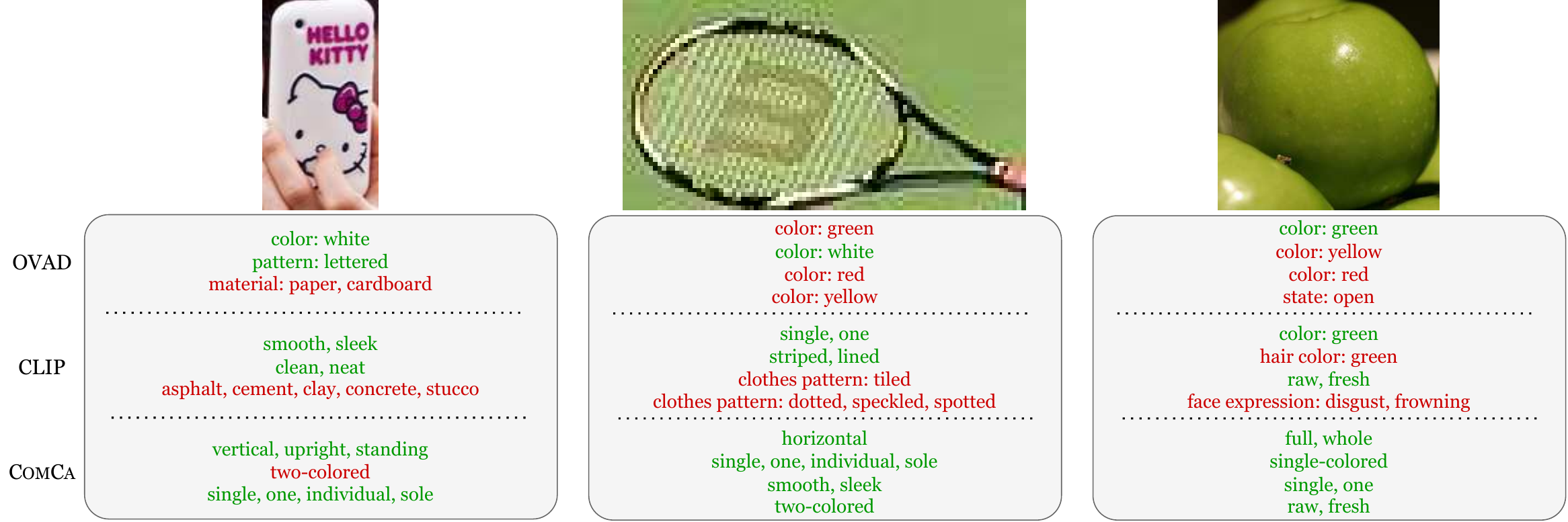}
        \caption{
            Comparison of performance on a mobile phone, a tennis racket, and an apple.
        }
        \label{fig:appendix:qualitative_results:1}
        \vspace{20pt}
    \end{subfigure}
    \begin{subfigure}[t]{\linewidth}
        \centering
        \includegraphics[width=\linewidth]{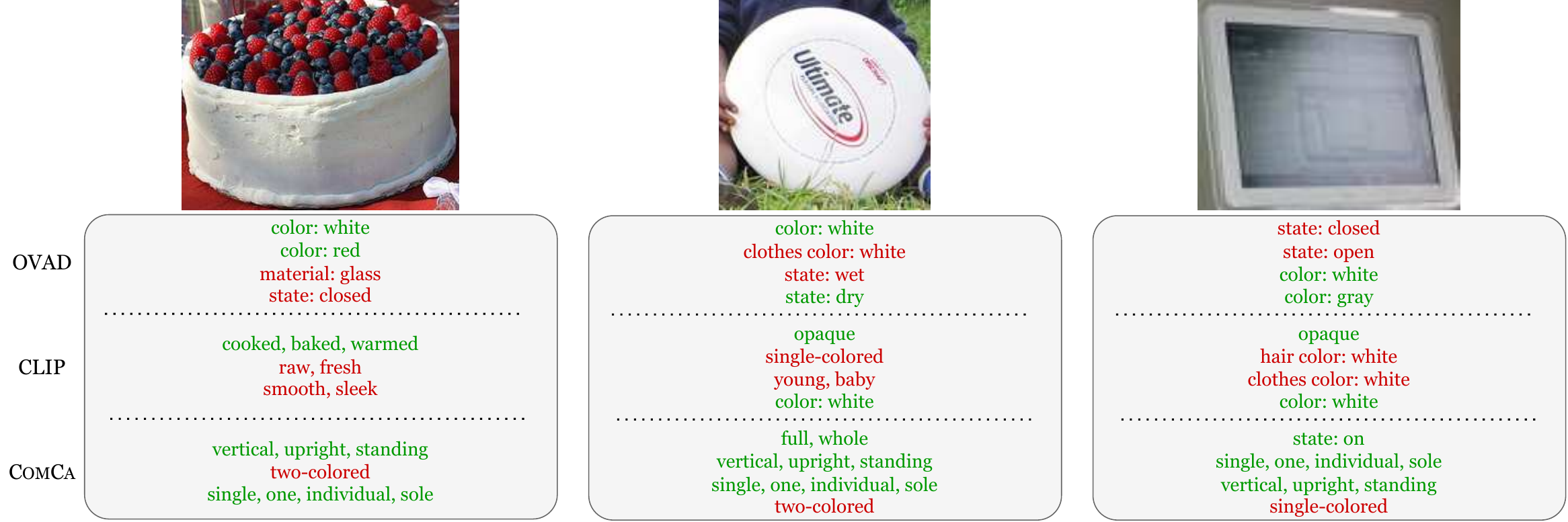}
        \caption{
            Comparison of performance on a cake, a frisbee flying disc, and a PC monitor.
        }
        \label{fig:appendix:qualitative_results:2}
        \vspace{20pt}
    \end{subfigure}
    \begin{subfigure}[t]{\linewidth}
        \centering
        \includegraphics[width=\linewidth]{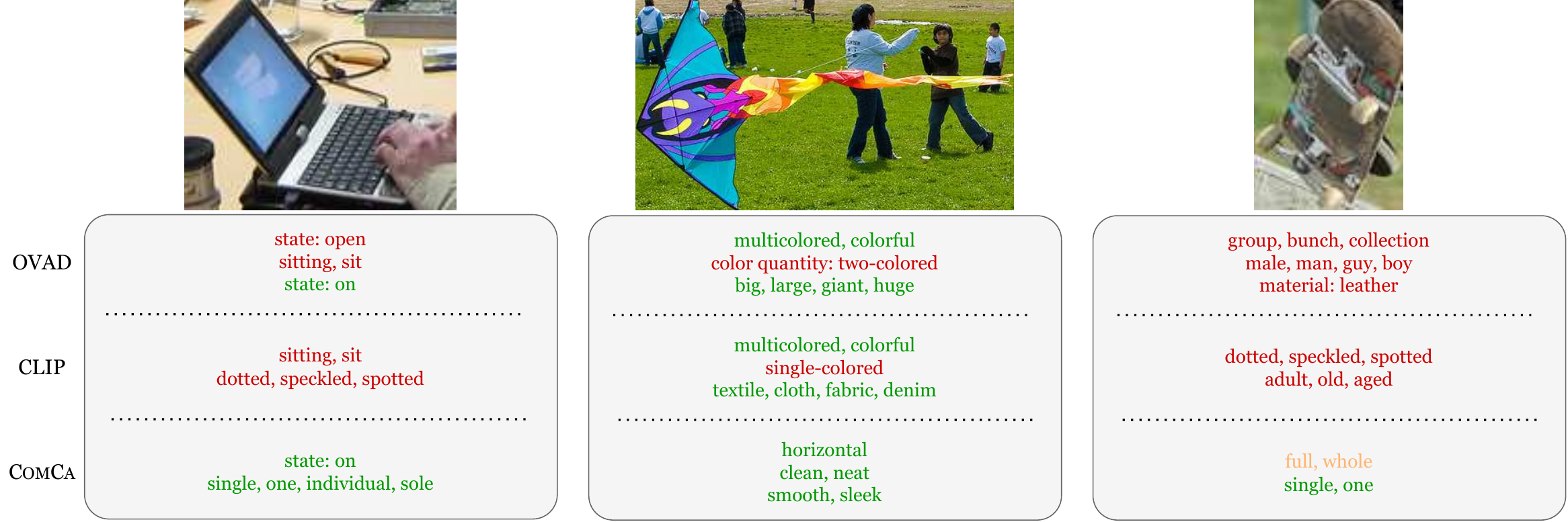}
        \caption{
            Comparison of performance on a laptop, a kite, and a skateboard.
        }
        \label{fig:appendix:qualitative_results:3}
    \end{subfigure}
    \caption{
        \textbf{Additional qualitative results.}
        Top positive predictions of OVAD, CLIP and \ours on sample images from OVAD.
        \inlineColorbox{DrawioGreen}{Green} are correct predictions, \inlineColorbox{DrawioRed}{red} are wrong ones.
    }
    \label{fig:appendix:qualitative_results}
    \vspace{20pt}
\end{figure*}

In \cref{fig:appendix:qualitative_results}, we provide additional qualitative results to those presented in Fig.~3 in \textit{Main}.
We compare OVAD and a vanilla zero-shot CLIP ViT-B/32 with \ours, based on the same architecture, similar to Fig.~3.
Firstly, we notice that OVAD often recognizes opposite attributes, in particular struggling with color.
For instance, in \cref{fig:appendix:qualitative_results:1} it detects ``green'', ``white'', ``red'', and ``yellow'' for the tennis racket and ``green'', ``yellow'', and ``red'' for the apple.
In addition, it struggles with materials, for example by predicting ``paper, cardboard'' for the mobile phone (\cref{fig:appendix:qualitative_results:1}) and ``leather'' for the skateboard (\cref{fig:appendix:qualitative_results:3}).
Similarly, we observe that also vanilla CLIP is prone to predicting contradictory attributes, such as ``cooked, baked, warmed'' and ``raw, fresh'' for the cake, or ``multicolored'' and ``single-colored'' for the kite.
Moreover, CLIP seems to disregard the part of the object an attribute is bound to: for instance, the apple is correctly predicted as ``green'', but also ``hair color'' is predicted as ``green''.
Similarly, the desktop monitor in \cref{fig:appendix:qualitative_results:2} is predicted as ``white'' three times: for ``color'' (correct), ``hair color'' (incorrect), and ``clothes color'' (incorrect).
On the other hand, \ours has a finer understanding of attributes and detects them more effectively, as demonstrated quantitatively in all our experiments.
Notably, it predicts ``two-colored'' for the mobile phone in the leftmost picture in \cref{fig:appendix:qualitative_results:1}: although it is wrong according to the ground truth, we argue that indeed it could be considered two-colored (white and pink).
Similarly, it predicts ``full, whole'' for the skateboard in \cref{fig:appendix:qualitative_results:3}: although that attribute is not annotated for that specific object (thus we mark it in yellow), we argue that it would be a correct prediction.

\section{Details on prompts}
\label{appendix:prompt_details}

We utilize five types of prompts: (i) for retrieval, to construct queries for web-scale datasets; (ii) for image generation; (iii) for soft labeling; (iv) for inference, to define prompts for the attributes of interest; and (v) to obtain attribute-object scores via LLMs.

\paragraph{Retrieval.} For retrieval, we construct prompts using the template \texttt{A photo of \{noun\} that is \{attribute\}}, following OvarNet~\cite{chen2023ovarnet}. For example, when retrieving images for a \texttt{red car}, the prompt will be \texttt{A photo of car that is red}.

\paragraph{Image generation.} To generate images we employ the following template:
\texttt{
a \{attr\} \{noun\} on a clear background, hyperrealistic, highly detailed, sharp focus, 4k
}
for attributes of type ``is'', as defined in OVAD~\cite{bravo2023ovad}, and the following template:
\texttt{
a \{noun\} with \{attr\} \{obj\} on a clear background, hyperrealistic, highly detailed, sharp focus, 4k
}
for attributes of type ``has''.

\paragraph{Soft labeling.} To compute soft labels, \ours first computes the similarity between the target attributes and the images in the cache. Following~\cite{radford2021clip},  we encapsulate attributes into templates to improve performance and reduce noise. We leverage the same templates of OVAD~\cite{bravo2023ovad}. For ``has'' type attributes, we use the following prompts:
\begin{itemize}
    \item \texttt{a \{attr\} \{obj\} \{noun\}}
    \item \texttt{a \{noun\} has \{attr\} \{obj\}}
\end{itemize}
For ``is'' type attributes, we use the following prompts:
\begin{itemize}
    \item \texttt{a \{attr\} \{noun\}}
    \item \texttt{a \{noun\} is \{attr\}}
\end{itemize}
As VAW~\cite{pham2021vaw} does not provide \textit{is}/\textit{has} type annotation, we use ``is'' type templates. Similarly to OVAD~\cite{bravo2023ovad}, we utilize the word ``object'' as the general noun to bind attributes to while computing these soft cache scores.

\paragraph{Inference.} Following OvarNet~\cite{chen2023ovarnet}, we construct prompts for inference by using the following template:
\textit{
A photo of something that is \{attribute\}
}.
For example, for the attribute ``clean'' the query produced at inference time will be \texttt{A photo of something that is clean}.

\paragraph{Object-attribute scoring via LLM.} When utilising LLMs to generate object-attribute compatibility scores we carefully instruct the model to ensure it provides sensible results. Specifically, we utilize the following template:\newline

\begin{tcolorbox}[breakable, enhanced jigsaw]
Let's play a role game. You will play the role of a researcher who is both a statistician and linguist. I will interpret a silly student who has many questions regarding language and statistics of language. \newline\newline
In particular, I will ask you to tell me which classes, or categories if you prefer, match or bind well with the attribute I will provide you. More precisely, you will have to tell me if each class/category that I will give you matches well the given attribute. You should also tell me how well they match on a scale 0 (the class cannot have the attribute) to 10 (the class can have the attribute and it is semantically fine to associate the attribute to the class). \newline\newline
Your response should list all the \{count\_categories\} classes, and provide for each one of them the score on the scale explained above. The output format should be `class: score'. No explanation at all, just plain output. \newline\newline
Additional rules:
\begin{itemize}
\item do not provide any outputs but the list of chosen categories
\item the output must be in the form of ``x. category: score'', where `x' is the index of the category
\item the output must be in the form of a list
\item make sure you provide a score for each category. There are \{count\_categories\} categories, so the output list must have \{count\_categories\} elements.
\end{itemize}
\leavevmode\newline
There are \{count\_categories\} classes (categories).
The list of classes, or categories, is the following:\newline
\{categories\} \newline\newline
The attribute is: \{attribute\}.
\end{tcolorbox}

Note that this template has different parameters:
\begin{description}
    \item[\texttt{\{count\_categories\}}] How many categories the prompt will contain.
    \item[\texttt{\{categories\}}] The list of categories to evaluate. More precisely, they are passed as a numbered list, with one category per line.
    \item[\texttt{\{attribute\}}] The attribute to which categories have to be bound for evaluation.
\end{description}

We empirically find that the LLM may struggle when handling hundreds or thousands of categories, as in the case of VAW~\cite{pham2021vaw}. Therefore, we find it useful to split the categories in multiple sets, construct independent prompts, and ask the model to respond to batches of categories separately, thus ensuring its context window is not filled.

\section{Resources used}
\label{appendix:resources-used}

Our training-free method does not require extensive computational resources and is designed to be computationally efficient.
We run all our experiments on a NVIDIA RTX 2080Ti GPU with 12GB of VRAM.
We use up to 4 NVIDIA RTX A6000 GPUs with 48GB of VRAM for the experiments that involve image generation, and are therefore more resource-demanding. Specifically, we use this configuration to generate images for the ablation in \cref{supp:table:ablation_gen_vs_ret}, which uses Stable Diffusion XL to generate images.
Evaluation time on the NVIDIA RTX 2080Ti GPU takes approximately 4 minutes on OVAD and 10 minutes on VAW.

\section{Limitations}
\label{appendix:limitations}

We presented the first approach for open-vocabulary training-free attribute detection. While effective, our method can be improved in several directions. First, \ours relies on an external database to generate the cache, so if this database does not contain a category, we cannot sample/bind it to an attribute. In some applications, using a general-purpose web-scale dataset might not suffice. 
Second, \ours relies on a large language model (LLM). If the LLM generates inaccurate or incorrect information, known as hallucinations, it could adversely affect the overall performance of the method.
Finally, our experimental evaluation demonstrates that \ours is competitive with training-based approaches and more effective when we aim to generalize to other datasets and domains (see Fig.~4 in \textit{Main}). However, training-based approaches may still be preferable if resources in terms of large scale datasets and computing infrastructure are available.

\end{document}